\documentclass{article}

 \usepackage[main, final]{neurips_2026}

\usepackage[utf8]{inputenc} 
\usepackage[T1]{fontenc}    
\usepackage{hyperref}       
\usepackage{url}            
\usepackage{booktabs}       
\usepackage{amsfonts}       
\usepackage{nicefrac}       
\usepackage{microtype}      
\usepackage{xcolor}         
\usepackage{pifont}
\usepackage{float}
\usepackage{stackengine}
\usepackage{enumitem}
\usepackage{float}

\usepackage{fontawesome5}
\definecolor{darkblue}{RGB}{178, 190, 201}
\definecolor{lightblue}{RGB}{218, 225, 231}
\definecolor{brown}{RGB}{242, 184, 160}
\definecolor{lightbrown}{RGB}{246,224,214}
\usepackage{subcaption}
\usepackage{tikz}
\usepackage{tabularx}
\usepackage{booktabs}
\usepackage{tikz}
\usepackage{xcolor}
\usepackage[table]{xcolor}
\newcommand{\partialcircle}{%
\tikz[baseline=-0.55ex]{
    \def\r{0.85ex} 
    \draw[orange!90!white,line width=0.85pt] (0,0) circle (\r);
    \fill[orange!90!white]
        (90:\r)
        arc (90:270:\r)
        -- (0,0) -- cycle;
}
}

\title{PhysVista: Benchmarking Physical Intelligence in VLMs via a Perception-Reasoning-Assessment Loop}

\author{
Xinge Peng$^{1}$  \;
Yiting Lu$^{1}$ \;
Tianwu Zhi$^{2}$ \;
Wen Wen$^{3}$ 
\;
Jianzhao Liu$^{2}$ \;
Xin Li$^{1}$
\thanks{Corresponding authors.}
\;
Zhibo Chen$^{1}$\,\footnotemark[1]
\\
$^{1}$University of Science and Technology of China \\
$^{2}$ByteDance \\
$^{3}$City University of Hong Kong \\
\texttt{xg.pengv@gmail.com}
}

\begin{document}

\maketitle

\vspace{-8.5mm}

\begin{center}
    \href{https://github.com/Helen1p/PhysVista}
    {\faGithub\ \textbf{Project}}
    \qquad
    \href{https://huggingface.co/datasets/HelenPeng/PhysVista}
    {\faDatabase\ \textbf{Dataset}}
\end{center}

\vspace{-1.0mm}
\begin{abstract}
\vspace{-3.5mm}
Vision-Language Models (VLMs) have shown strong multimodal reasoning capabilities, yet whether they truly capture the physical consistency underlying real-world dynamics remains unclear. Existing benchmark paradigms often suffer from fragmented evaluation, focusing on isolated cognitive stages while overlooking the inherent synergy between perception, reasoning, and physical judgment. The lack of a holistic perspective limits the ability to diagnose whether VLMs can reliably serve as evaluators of physical plausibility in generative content. 
To address these issues, we introduce \textbf{PhysVista}, a benchmark designed to evaluate physical intelligence in VLMs through a closed cognitive loop framework inspired by the human \textbf{seeing–reasoning–assessment} process. PhysVista restores this loop by jointly evaluating physical state perception, physical dynamics reasoning, and physical plausibility assessment. It further distinguishes \textbf{event-level} reasoning and \textbf{scale-level} reasoning to enable fine-grained analysis of physical understanding. In addition, PhysVista incorporates both real-world and AI-generated videos, allowing evaluation across diverse domains and emerging generative scenarios. 
Extensive experiments across a diverse set of VLMs reveal substantial limitations in physical reasoning and plausibility assessment, highlighting a persistent gap between visual recognition and genuine physical understanding, and pointing toward more principled designs for physically grounded multimodal intelligence.
\end{abstract}

\vspace{-7mm}

\begin{figure}[H]
  \centering
\includegraphics[width=0.9\linewidth]{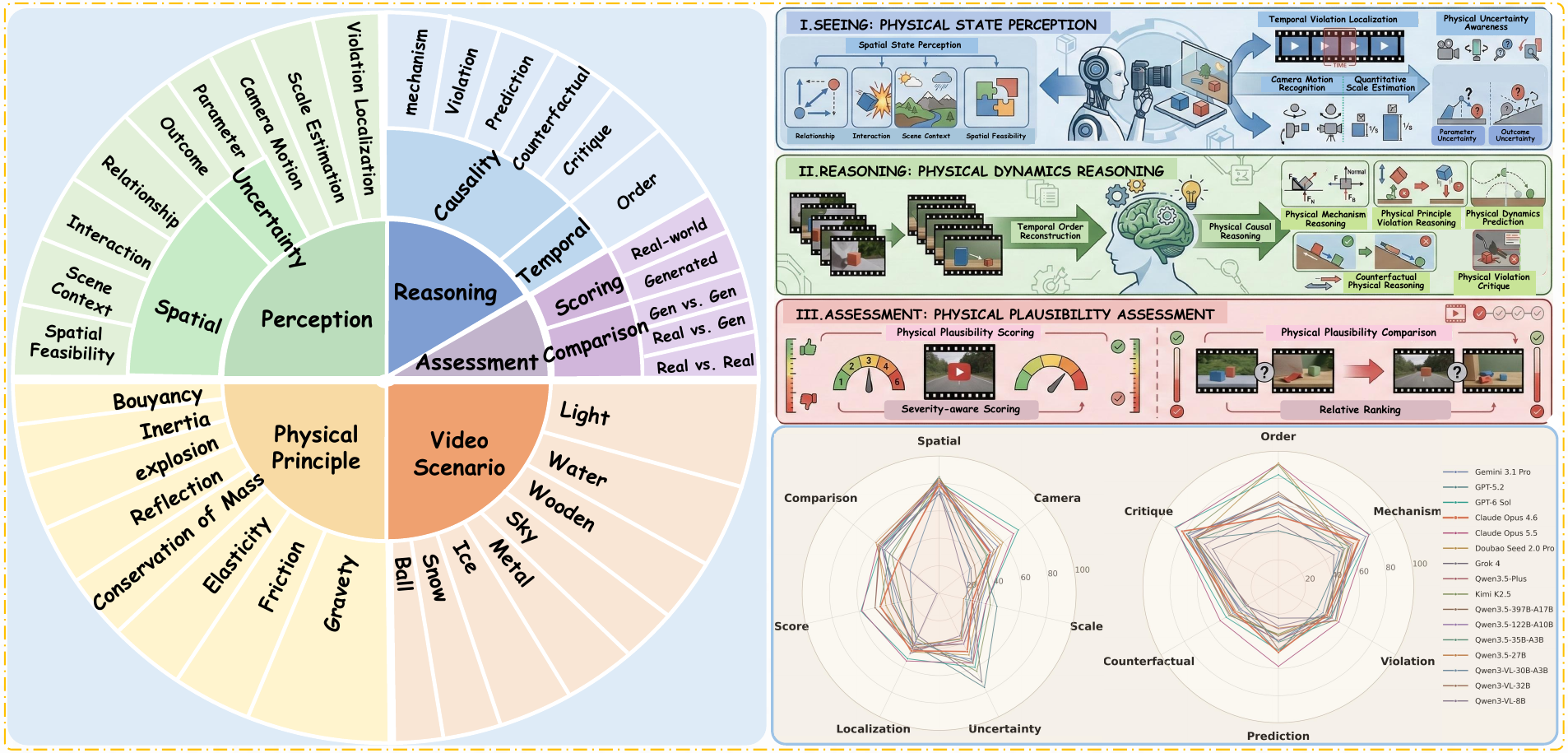}
  \vspace{-1.5mm}
  \caption{Overview of PhysVista. PhysVista evaluates physical intelligence in VLMs through \emph{Seeing}, \emph{Reasoning}, and \emph{Assessment} across real and generated videos.
  }
  \label{fig:abs}
\end{figure}

\section{Introduction}
\label{sec:intro}
\vspace{-3.5mm}
Vision-Language Models (VLMs)~\cite{guo2025deepseek,wang2025internvl3,Qwen3-VL,team2026kimi,abdin2024phi} have achieved remarkable progress in visual understanding, reasoning, and multimodal interaction. 
Driven by their strong performance, some works have increasingly explored VLMs for \textit{visual quality assessment}~\cite{zhang2024q,wen2026self,lu2025omniquality,peng2026iqa}, with recent efforts going beyond \textit{appearance-level evaluation} to judge the \textit{physical plausibility} of content produced by generative models~\cite{lu20254dworldbench,meng2024towards}. 
However, the reliability of such physical judgments remains unclear, raising a fundamental question: do VLMs truly understand the underlying physical dynamics of the world? 
Recent studies~\cite{mak2026physicsmind,wangphysunibench,shen2025phyx} reveal that modern models frequently produce physically inconsistent interpretations, suggesting that strong visual recognition does not necessarily translate into genuine physical understanding. 
However, the extent of this limitation remains unclear due to the lack of systematic evaluation of physical intelligence in VLMs. 
\vspace{-1mm}

\textbf{Physical understanding inherently follows a closed cognitive loop.}
Humans interpret physical events by first perceiving observable states, then reasoning about the underlying causal mechanisms, and finally assessing whether the observed dynamics conform to physical laws. 
This \emph{seeing–reasoning–assessment} loop forms the foundation of physical intelligence, enabling consistent interpretation, prediction, and validation of dynamic environments. 
However, existing benchmarks evaluate physical understanding only at fragmented stages of this cognitive loop. 
PhysBench~\cite{chow2025physbench} primarily focuses on perception-level evaluation. 
In terms of reasoning, most physical reasoning benchmarks~\cite{foss2025causalvqa,riochet2018intphys,krojer2025shortcut,weicophybench} cover only a limited range of tasks and evaluation capabilities. 
Moreover, they mainly evaluate reasoning at the \emph{event-level}, focusing on whether an event occurs, rather than \emph{scale-level} reasoning that requires distinguishing fine-grained differences in magnitude, quantity, precise position, or proportion. 
The most recent work, PAI-Bench~\cite{zhou2025pai}, evaluates both perceptual and reasoning aspects of physical understanding. 
However, it still lacks the final component required to complete the closed loop—\emph{assessment}. 
A comparison of task types is illustrated in Tab.~\ref{tab:PhysVista_comparison}. 
As a result, current benchmarks measure isolated manifestations of physical intelligence rather than holistic physical cognition. 

\vspace{-1mm}

\textbf{Beyond limited task coverage, the evaluation paradigm itself remains problematic.}
PhysBench~\cite{chow2025physbench} suffers from language bias that introduces shortcuts, allowing models to answer questions in a video-blind manner. 
IntPhys2~\cite{bordes2025intphys} formulates evaluation as binary plausibility judgments of events, which mainly tests shallow physical intuition rather than deeper causal reasoning. 
Other benchmarks~\cite{mak2026physicsmind,wangphysunibench,shen2025phyx} focus on applying physical laws in structured problem-solving settings, where visual information serves only as auxiliary cues. 
This setup is fundamentally misaligned with the real-world physical understanding, which is perception-driven and must be grounded in visual observations. 
Consequently, current benchmarks fall short of evaluating comprehensive physical understanding in a reliable, in-depth way. 

\vspace{-1mm}

\textbf{Additionally, the lack of domain diversity in data source also limits the scope of evaluation.}
Early benchmarks~\cite{jassim2023grasp,riochet2018intphys,weihs2022benchmarking,tung2023physion++,ates2022craft,baradel2019cophy,yi2019clevrer,rajani2020esprit} are predominantly simulation-based, with limited evaluation on real-world videos. 
However, the simplicity of simulated environments limits their ability to faithfully reflect a model’s capability in reasoning about complex physical dynamics in real-world scenarios. 
Recent works~\cite{foss2025causalvqa,zhou2025pai,gundawar2025pac} have begun incorporating real-world videos into evaluation. 
Meanwhile, the rapid emergence of generative video models~\cite{wu2025hunyuanvideo,chen2024videocrafter2} and the widespread use of AI-generated videos make the evaluation of generated content increasingly important, yet this aspect remains largely unexplored. 
Generated videos often exhibit physical implausibilities arising from complex scene changes, object interactions, and human motions. 
However, the gap in VLM physical understanding between real-world and generated videos remains unclear. 
Moreover, such content introduces new evaluation requirements, including tasks such as Physical Violation Critique and Physical Plausibility Scoring.

To address these challenges, we introduce \textbf{PhysVista}, a benchmark for evaluating physical intelligence in Vision-Language Models through a unified closed-loop framework. 
Rather than emphasizing new data collection, PhysVista focuses on principled task formulation, systematic re-annotation, and unified evaluation for physical understanding. 
It systematically measures physical state perception, temporal and causal reasoning, and plausibility assessment, enabling holistic diagnosis beyond isolated capability testing. 
It also includes both real-world and AI-generated videos. 
Beyond task and data diversity, we further design an elaborate evaluation framework that assesses reasoning at both the event and scale levels. 
Our main contributions are as follows:
\begin{itemize}[
leftmargin=2em,
labelindent=0pt,
labelsep=0.5em,
itemindent=0pt,
itemsep=0.3em,
topsep=0pt,
parsep=0pt
]
\item \textbf{PhysVista Benchmark.} 
We introduce \textbf{PhysVista}, a benchmark for evaluating physical intelligence in Vision-Language Models that restores the \emph{seeing–reasoning–assessment} cognitive loop. 
It covers perception, reasoning, and plausibility evaluation using both real-world and AI-generated videos.
\item \textbf{Versatile Evaluation Framework.}
We propose a versatile evaluation framework that measures physical state perception, temporal and causal reasoning, and plausibility assessment. 
It further distinguishes \emph{event-level} and \emph{scale-level} reasoning to enable fine-grained evaluation of physical understanding.
\item \textbf{Key Insights into VLM Physical Intelligence.}
We provide detailed performance analysis and uncover important observations that highlight the limitations of current VLMs, as well as inform the design of future physically grounded models.
\end{itemize}

\vspace{-2.5mm}
\section{Related Work}
\vspace{-2mm}
\subsection{Physics-related Benchmarks for Generative Models.}
\vspace{-1mm}
While traditional metrics (\textit{e.g.}, FVD~\cite{ge2024content}) and multi-dimensional benchmarks~\cite{huang2024vbench,zheng2025vbench,sun2025t2v,duan2025worldscore} assess visual and fine-grained attributes, they neglect long-term physical modeling. To address this, recent VLM-based benchmarks explicitly evaluate physical dynamics and causality: PhyGenBench~\cite{meng2024towards} and WorldBench~\cite{upadhyay2026worldbench} analyze rule decomposition, PhyWorldBench~\cite{gu2025phyworldbench} tests anti-physics scenarios, and VideoVerse~\cite{wang2025videoverse} assesses causal QA. For closed-loop decision-making, DrivingGen~\cite{zhou2026drivinggen} tests driving trajectory safety, WorldArena~\cite{shang2026worldarena} simulates environments, and 4DWorldBench~\cite{lu20254dworldbench} evaluates 3D/4D spatiotemporal consistency. However, these methods often output single scalars without structural decomposition. Relying solely on final generations without continuous state supervision also obscures long-term error propagation, highlighting the critical need for systematic, interactive evaluation frameworks.
\vspace{-2mm}
\subsection{Physics-related Benchmarks for VLMs.}
Early benchmarks~\cite{jassim2023grasp,riochet2018intphys,weihs2022benchmarking,tung2023physion++,ates2022craft,baradel2019cophy,yi2019clevrer,rajani2020esprit,bordes2025intphys} adopt simulation-based environments to assess intuitive physics through violation-of-expectation paradigms, they often lack the visual complexity and distributional diversity of real-world or model-generated videos. 
PhysBench~\cite{chow2025physbench} emphasizes physical perception by covering object properties, spatial relations, scene dynamics, and motion understanding, but suffers from language shortcut. 
To overcome this, MVP Bench~\cite{krojer2025shortcut} construct controlled minimal video pairs to reduce shortcut learning and evaluate models’ sensitivity to physically inconsistent events. 
Another line of works primarily evaluate the causal reasoning ability in real-world videos. 
CausalVQA~\cite{foss2025causalvqa} and CoPhyBench~\cite{weicophybench} focus on predicting outcomes and reasoning under interventions or conditional observations, probing models’ ability to infer causal physical mechanisms. 
Other physical benchmarks~\cite{mak2026physicsmind,wangphysunibench,shen2025phyx} focus on structured physical problem-solving in exam-style settings, which fails to reflect open-world physical understanding. 
There are also some application-oriented evaluation benchmarks~\cite{zhou2025pai,gundawar2025pac} targets for assessing physical understanding particularly in embodied AI and robotics scenarios. 
More recently, ExeWorldBench~\cite{peng2026exeworldbench} has integrated physical reasoning with 3D coding. 
Overall, existing benchmarks focus on isolated aspects of physical understanding while largely neglecting evaluation on AI-generated content. 

\begin{table*}[t]
\centering
\small
\setlength{\tabcolsep}{4pt}
\renewcommand{\arraystretch}{1.15}

\resizebox{\textwidth}{!}{
\begin{tabular}{l l c c c c c c c c c}
\toprule

& &
\multicolumn{5}{c}{\textbf{Perception}} &
\multicolumn{2}{c}{\textbf{Reasoning}} &
\multicolumn{2}{c}{\textbf{Assessment}} \\

\cmidrule(lr){3-7}
\cmidrule(lr){8-9}
\cmidrule(lr){10-11}

\textbf{Benchmark} &
\textbf{Data Source} &
\textbf{Spatial} &
\textbf{Localization} &
\textbf{Camera} &
\textbf{Scale} &
\textbf{Uncertainty} &
\textbf{Order} &
\textbf{Causality} &
\textbf{Scoring} &
\textbf{Comparison} \\

\midrule

IntPhys2~\cite{bordes2025intphys} &
Simulation &
\textcolor{red}{\ding{55}} &
\textcolor{red}{\ding{55}} &
\textcolor{red}{\ding{55}} &
\textcolor{red}{\ding{55}} &
\textcolor{red}{\ding{55}} &
\textcolor{red}{\ding{55}} &
\partialcircle &
\textcolor{red}{\ding{55}} &
\textcolor{red}{\ding{55}} \\

PhysBench~\cite{chow2025physbench} &
Real+Simulation &
\textcolor{green!60!black}{\checkmark} &
\textcolor{red}{\ding{55}} &
\partialcircle &
\textcolor{red}{\ding{55}} &
\textcolor{red}{\ding{55}} &
\textcolor{green!60!black}{\checkmark} &
\partialcircle &
\textcolor{red}{\ding{55}} &
\textcolor{red}{\ding{55}} \\

CausalVQA~\cite{foss2025causalvqa} &
Real &
\partialcircle &
\textcolor{red}{\ding{55}} &
\textcolor{red}{\ding{55}} &
\textcolor{red}{\ding{55}} &
\textcolor{red}{\ding{55}} &
\textcolor{red}{\ding{55}} &
\partialcircle &
\textcolor{red}{\ding{55}} &
\textcolor{red}{\ding{55}} \\

MVP Bench~\cite{krojer2025shortcut} &
Real+Embodied+Simulation &
\partialcircle &
\textcolor{red}{\ding{55}} &
\textcolor{red}{\ding{55}} &
\textcolor{red}{\ding{55}} &
\textcolor{red}{\ding{55}} &
\textcolor{red}{\ding{55}} &
\partialcircle &
\textcolor{red}{\ding{55}} &
\textcolor{red}{\ding{55}} \\

PAI-Bench~\cite{zhou2025pai} &
Real+Embodied &
\textcolor{green!60!black}{\checkmark} &
\textcolor{red}{\ding{55}} &
\partialcircle &
\textcolor{red}{\ding{55}} &
\textcolor{red}{\ding{55}} &
\textcolor{green!60!black}{\checkmark} &
\partialcircle &
\textcolor{red}{\ding{55}} &
\textcolor{red}{\ding{55}} \\

\midrule
\rowcolor{gray!15}
\textbf{PhysVista (Ours)} &
Real + Generated Videos &
\textcolor{green!60!black}{\checkmark} &
\textcolor{green!60!black}{\checkmark} &
\textcolor{green!60!black}{\checkmark} &
\textcolor{green!60!black}{\checkmark} &
\textcolor{green!60!black}{\checkmark} &
\textcolor{green!60!black}{\checkmark} &
\textcolor{green!60!black}{\checkmark} &
\textcolor{green!60!black}{\checkmark} &
\textcolor{green!60!black}{\checkmark} \\

\bottomrule
\end{tabular}
}

\caption{
Comparison with related physical understanding benchmarks. PhysVista provides unified evaluation across perception, reasoning, and assessment, and explicitly includes generated-video scenarios. 
}

\label{tab:PhysVista_comparison}
\end{table*}

\section{PhysVista}

\subsection{Overview}
As shown in Fig.~\ref{fig:abs}, PhysVista is a benchmark designed to evaluate physical intelligence in Vision-Language Models through a closed-loop cognitive framework consisting of three stages: Physical State Perception (Seeing), Physical Causal Reasoning (Reasoning), and Physical Plausibility Assessment (Assessment). 
Unlike prior benchmarks that evaluate isolated physical capabilities, PhysVista systematically measures whether models can reconstruct observable physical states, infer causal mechanisms, and assess global physical Plausibility with scores and ranks.
The benchmark incorporates both real-world and AI-generated videos containing diverse object interactions, motion dynamics, and varying degrees of physical consistency, enabling holistic evaluation of physical cognition beyond event-level detection. 
More detailed benchmark statistics are provided in Appendix.

\vspace{-1.5mm}
\subsection{Seeing: Physical State Perception}
\subsubsection{Task Information.}
The Physical State Perception stage evaluates whether models can accurately recover observable physical states directly from visual inputs. 
We consider five complementary aspects: spatial state perception, temporal state perception, camera motion recognition, quantitative scale estimation, and physical uncertainty awareness.

\noindent \textsc{\textbf{Spatial State Perception.}}
We decompose spatial perception into four subtypes reflecting different dimensions of spatial cognition. 
\begin{itemize}[
leftmargin=2em,
labelindent=0pt,
labelsep=0.5em,
itemindent=0pt,
itemsep=0.3em,
topsep=0pt,
parsep=0pt
]
\item \textit{\textsc{\textbf{Relationship.}}}
Evaluates whether models can correctly identify relative spatial configurations among objects, including direction and distance.~(\textit{e.g.}, left/right, near/far, containment, support).
\item \textit{\textsc{\textbf{Interaction.}}}
Examines recognition of physically meaningful interactions between entities, such as contact, collision.
\item \textit{\textsc{\textbf{Scene Context.}}}
Assesses whether models can perceive environmental conditions that influence physical interpretation, including terrain type, weather, or surface properties observable from the scene.
\item \textit{\textsc{\textbf{Spatial Feasibility.}}}
Measures the ability to judge whether spatial configurations are geometrically and physically realizable under real-world constraints (\textit{e.g.}, size compatibility or object fitting).
\end{itemize}
\noindent \textsc{\textbf{Temporal Violation Localization.}}
This task requires models to identify the precise time interval during which a physically inconsistent event occurs. 
Unlike global video-level plausibility detection in prior works, this task requires second-level fine-grained temporal grounding of specific physical violations. 

\noindent \textsc{\textbf{Camera Motion Recognition.}}
Although prior works have explored this task, they lack a systematic taxonomy of camera motions and cover only limited categories, leaving camera motion recognition largely underexplored. The model is required to classify camera motion into ten predefined categories given a video.

\noindent \textsc{\textbf{Quantitative Scale Estimation.}}
While existing benchmarks rely on coarse descriptive adjectives to assess relative scale perception, we reformulate the task in a quantitative manner by requiring explicit relative ratio estimation between objects and express responses as fractional values (\textit{e.g.}, $1/3$), enabling fine-grained and numerically grounded physical perception.

\noindent \textsc{\textbf{Physical Uncertainty Awareness.}}
We incorporate uncertainty awareness as a fundamental dimension of physical perception, which remains largely neglected in existing benchmarks. 
We define two complementary forms of uncertainty: 
\begin{itemize}[
leftmargin=2em,
labelindent=0pt,
labelsep=0.5em,
itemindent=0pt,
itemsep=0.3em,
topsep=0pt,
parsep=0pt
]
\item \textit{\textsc{\textbf{Parameter Uncertainty Awareness.}}}
The ability to recognize when latent physical parameters (\textit{e.g.}, friction coefficients or reflectance properties) cannot be determined from visual evidence alone. 
\item \textit{\textsc{\textbf{Outcome Uncertainty Awareness.}}}
Assesses whether models correctly identify scenarios in which future physical outcomes are inherently indeterminate due to incomplete information or long-term, unobservable processes. 
\end{itemize}
Together, these tasks establish the perceptual foundation required for subsequent physical reasoning and assessment.
\subsubsection{Data Collection.}

Our data collection follows two stages: \textbf{video curation} and \textbf{question--answer construction}.
\vspace{-2mm}
\paragraph{Stage I: Video curation.}
Except for \textbf{Camera Motion Recognition}, videos are sourced from the real-world dataset WISA-80K~\cite{wang2025wisa} and the video-generation dataset VideoPhy-2~\cite{bansal2025videophy}, both covering diverse physical categories for physical understanding.
For \textbf{Camera Motion Recognition}, we use Multi-CamVideo~\cite{bai2025recammaster}, which provides accurate camera pose trajectories. We select clips with a \emph{single dominant} camera motion and no interfering movements to enable unambiguous motion categorization.
To ensure sufficient task complexity, we additionally remove overly simple scenes that lack rich multi-object interactions or meaningful physical dynamics.
All tasks are formulated as \textbf{single-answer multiple-choice} questions.
\vspace{-2mm}
\paragraph{Stage II: Question--answer construction.}
For each task, we manually design a \textbf{single fixed template} (\textit{e.g.}, \textbf{Spatial State Perception}, \textbf{Temporal Order Reconstruction}, and \textbf{Camera Motion Recognition}). Each template is task-specific but \emph{content-agnostic} (no slot filling or instance-dependent variables), ensuring consistent structure and minimizing auxiliary textual cues.
\vspace{-2mm}
\paragraph{Task-specific option construction.}
\textbf{Spatial State Perception:} options are generated from sampled frames and the original dataset captions.
\textbf{Temporal Violation Localization:} human annotators label the start and end timestamps of intervals exhibiting physically implausible behavior.
\textbf{Camera Motion Recognition:} we automatically assign one of ten predefined motion types using camera pose trajectories. Motions are first grouped into rotation-only (\emph{e.g.}, pan/tilt) versus translation-with-rotation based on camera centers and viewing directions; the latter are further split into arc motions and pure translational motions by predefined geometric criteria.
\textbf{Quantitative Scale Estimation:} human annotators provide explicit relative object ratios using fractional values as precise ground truth.
\textbf{Physical Uncertainty Awareness:} for \textbf{Parameter Uncertainty}, we prompt Gemini 3.1 Pro~\cite{Gemini} to generate questions about latent physical parameters; for \textbf{Outcome Uncertainty}, questions follow VideoPhy-2 indeterminate-outcome annotations, where the correct answer is deterministically \emph{Indeterminate}.
\vspace{-2mm}
\paragraph{Human verification.}
We conduct human verification for tasks whose ground truth is produced using VLMs (\textit{e.g.}, \textbf{Spatial State Perception} and \textbf{Parameter Uncertainty}) to ensure data quality. Additionally, we also employ human annotators to rewrite samples exhibiting strong stylistic patterns, thereby reducing potential language bias. 
\subsection{Reasoning: Physical Dynamics Reasoning}

\begin{figure}[tb]
  \centering
\includegraphics[width=1.0\linewidth]{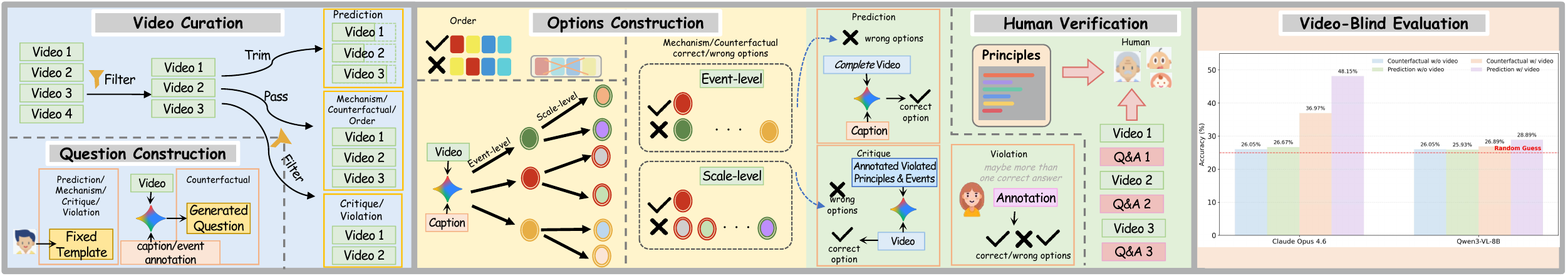}
\vspace{-3mm}
  \caption{Data collection \& evaluation of Physical Dynamics Reasoning. 
  }
  \label{fig:data_collection}
\end{figure}

\subsubsection{Task Information.}
The Physical Dynamics Reasoning stage evaluates whether models can reason about underlying physical processes beyond directly observable appearances. 
Unlike perception-level tasks that focus on state recognition, this stage requires inferring temporal dependencies and several causal mechanisms with physical principles. 
We decompose this stage into two complementary reasoning dimensions: temporal reasoning and causal reasoning, instantiated by Temporal Order Reconstruction and five causal reasoning tasks, respectively. 

\noindent \textsc{\textbf{Temporal Order Reconstruction.}}
Given temporally shuffled video frames sampled at irregular intervals, the model reconstructs a coherent physical event sequence. This task evaluates physical temporal reasoning by testing whether models can infer physically grounded temporal dependencies in real-world videos while remaining robust to minor inconsistencies in generated content.

\noindent \textsc{\textbf{Physical Causal Reasoning.}}
Beyond temporal dependencies, this stage evaluates comprehensive physical causal reasoning through five complementary subtasks:
\begin{itemize}[
leftmargin=2em,
labelindent=0pt,
labelsep=0.5em,
itemindent=0pt,
itemsep=0.3em,
topsep=0pt,
parsep=0pt
]
\item \textit{\textsc{\textbf{Physical Mechanism Reasoning.}}}
Requires inferring the causal physical mechanism governing the observed event, rather than providing a superficial description of its appearance.
\item \textit{\textsc{\textbf{Physical Principle Violation Reasoning.}}}
Requires determining all instances of violations of fundamental physical principles (\textit{e.g.}, conservation laws, gravity) present in the video. 
\item \textit{\textsc{\textbf{Physical Dynamics Prediction.}}}
Assesses whether models can predict subsequent physical states or outcomes based on current observations and inferred dynamics, requiring event-level specificity and fine-grained scale-level distinctions (\textit{e.g.}, stopping before versus exactly at a reference point).
\item \textit{\textsc{\textbf{Counterfactual Physical Reasoning.}}}
Evaluates the ability to reason about hypothetical scenarios by modifying a key physical condition (\textit{e.g.}, surface friction or applied force) and predicting the resulting outcome with event-level and scale-level precision. 
\item \textit{\textsc{\textbf{Physical Violation Critique.}}}
Requires models to identify physically implausible events, analyze the underlying violation of real-world physical constraints, and provide a principled critique by suggesting modifications to the key physical factors that would render the scenario physically plausible.
\end{itemize}
\subsubsection{Data Collection.}
As illustrated in Fig.~\ref{fig:data_collection}, our \textbf{Physical Causal Reasoning} data collection contains two stages: \textbf{video curation} and \textbf{question--answer construction}.
\vspace{-2mm}
\paragraph{Stage I: Video curation.}
We first filter out static or low-dynamic clips and retain videos with rich physical motion and multi-object interactions. We then apply task-specific filtering to ensure unambiguous evaluation:
(i) for \textbf{Physical Violation Critique}, we exclude videos containing multiple simultaneous principle violations;
(ii) for \textbf{Physical Principle Violation Reasoning}, we remove intermediate or ambiguous violations to keep the violated principle clear and consistent;
(iii) for \textbf{Physical Dynamics Prediction}, we only provide the early segment as input and withhold the final outcome event.
All tasks are formulated as single-answer multiple-choice questions, except \textbf{Physical Principle Violation Reasoning}.
\vspace{-2mm}
\paragraph{Stage II: Question--answer construction.}
For all tasks except \textbf{Counterfactual Physical Reasoning}, we use a \textbf{fixed, content-agnostic template} that is invariant across instances to minimize linguistic cues and video-blind answering. For \textbf{Counterfactual Physical Reasoning}, questions are centered on the primary physical event. We prompt Gemini 3.1 Pro to identify the main event using sampled frames and captions or event annotations from WISA-80K~\cite{wang2025wisa} and VideoPhy-2~\cite{bansal2025videophy}, respectively. Counterfactual conditions are introduced \emph{implicitly} by modifying event scenarios (\textit{e.g.}, replacing an icy surface with a cobbled road) rather than explicitly stating parameter changes.
\vspace{-2mm}
\paragraph{Option design principles.}
Answer generation is task-specific but follows four shared principles:
(1) all options are physically plausible and consistent with basic intuition;
(2) all options align with the video context;
(3) the correct option has clearly superior causal explanatory power with minimal ambiguity; and
(4) incorrect options are comparable to the correct one in either event-level structure or scale-level granularity.
\vspace{-2mm}
\paragraph{Task-specific option construction.}
For \textbf{Temporal Order Reconstruction}, four frames are sampled at unequal temporal intervals, and distractors are constructed by swapping adjacent frames only. 
For \textbf{Physical Mechanism Reasoning} and \textbf{Counterfactual Physical Reasoning}, we adopt a hierarchical three-stage procedure:
\emph{(1) Event-level generation:} produce and rank four coarse-grained base options, selecting the top-ranked as the candidate correct answer;
\emph{(2) Scale-level refinement:} refine each base option into semantically similar variants with fine-grained distinctions and rank them within each base group;
\emph{(3) Selection:} use (a) \emph{coarse-grained selection} by taking the top-ranked option from each base group to evaluate event-level reasoning, and (b) \emph{fine-grained selection} by sampling multiple variants within the same base group (optionally alongside other groups) to test scale sensitivity. 
For \textbf{Physical Principle Violation Reasoning}, correct answers come from annotated violated principles, while distractors are sampled from non-violated principles within the same context, potentially yielding multiple correct answers.
For \textbf{Physical Dynamics Prediction}, distractors follow the same hierarchical strategy and the correct option is derived from the withheld subsequent segment.
For \textbf{Physical Violation Critique}, we prompt Gemini 3.1 Pro to minimally edit the annotated violated principles and associated events such that the revised scenario becomes physically coherent; this annotation-grounded prompting constrains reasoning and mitigates hallucinations. Distractors include both event-level and scale-level alternatives.
\vspace{-2mm}
\paragraph{Human verification.}
Finally, we conduct human verification to: (i) ensure data quality. 
(ii) Remove samples that can be answered correctly without viewing the video. 
The video-blind evaluation in Fig.~\ref{fig:data_collection} further demonstrates that the dataset is unbiased. 
(iii) Paraphrase samples with recognizable LLM-style phrasing to minimize potential language bias. 
\subsection{Assessment: Physical Plausibility Assessment}
\subsubsection{Task Information.}
The Physical Plausibility Assessment stage measures a model’s comprehensive understanding of physical plausibility by requiring both 5-level scoring for single videos and relative plausibility comparison between video pairs. 
Compared to perception and reasoning stages that focus on recognizing physical states or inferring causal mechanisms, the assessment stage emphasizes holistic evaluation of physical plausibility across the entire video, particularly for generated videos where subtle physical inconsistencies may arise, requiring models to calibrate their judgments to the \emph{severity} of such inconsistencies. 

\noindent \textsc{\textbf{Physical Plausibility Scoring.}}
Requires assigning a discrete plausibility rating from 1 to 5 to a single video by evaluating its overall physical coherence, with the score calibrated to the severity of physical inconsistencies rather than a binary valid/invalid decision.

\noindent \textsc{\textbf{Physical Plausibility Comparison.}}
Requires determining which video exhibits higher physical plausibility when comparing two samples, based on their overall physical coherence and realism. 
This task provides a relative judgment setting that is less sensitive to score-scale evaluation and directly tests whether the model can rank physical realism between samples. 
\subsubsection{Data Collection.}
Our data collection consists of two stages: \textbf{video curation} and \textbf{question–answer construction}. 
\vspace{-2mm}
\paragraph{Stage I: Video curation.}
We draw generated videos from VideoPhy-2~\cite{bansal2025videophy} and real-world videos from WISA-80K~\cite{wang2025wisa}. 
VideoPhy-2 provides physical plausibility annotations with discrete scores ranging from 1 to 5. 
To ensure balanced coverage of different realism levels, we uniformly sample generated videos across all five score categories. 
In addition, real-world videos from WISA-80K are incorporated as physically consistent samples and assigned the highest plausibility score of 5. 
\vspace{-2mm}
\paragraph{Stage II: Question--answer construction.}
We adopt fixed templates of question and answer for both scoring and comparison tasks, and all answers are formatted as multiple-choice options. 
For the Physical Plausibility Comparison task, we construct three types of video pairs: real-world vs.\ generated, real-world vs.\ real-world, and generated vs.\ generated. 
In the generated–generated setting, which constitutes the most fine-grained comparison among three types, pairs are constructed to control for semantic similarity, and most pairs differ by no more than two plausibility levels, encouraging models to distinguish subtle differences in physical coherence.
\vspace{-1mm}
\section{Experiments}
\vspace{-1mm}
Following the design of PhysVista, experiments are organized into three stages corresponding to physical cognition: Physical State Perception (Seeing), Physical Causal Reasoning (Reasoning), and Physical Plausibility Assessment (Assessment). 
This evaluation protocol enables comprehensive diagnosis of model capabilities from observable state perception to causal inference and global physical judgment.

\begin{figure}[tb]
  \centering
  \begin{subfigure}[c]{0.3\linewidth}
    \centering
    \vfill
    \includegraphics[width=\linewidth]{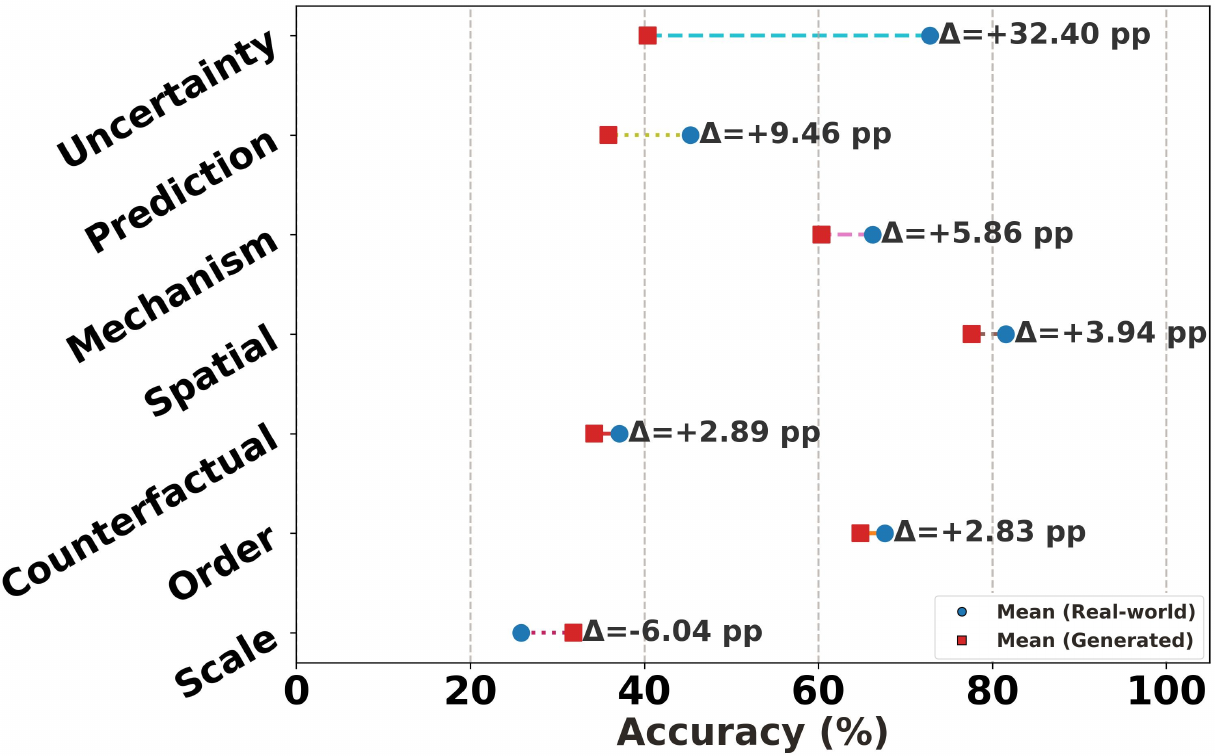}
    \vfill
    \caption{Performance gap between on real-world and generated videos across tasks.}
    \label{fig:gap}
  \end{subfigure}
  \hfill
  \begin{subfigure}[c]{0.32\linewidth}
    \centering
    \vfill
    \includegraphics[width=0.9\linewidth]{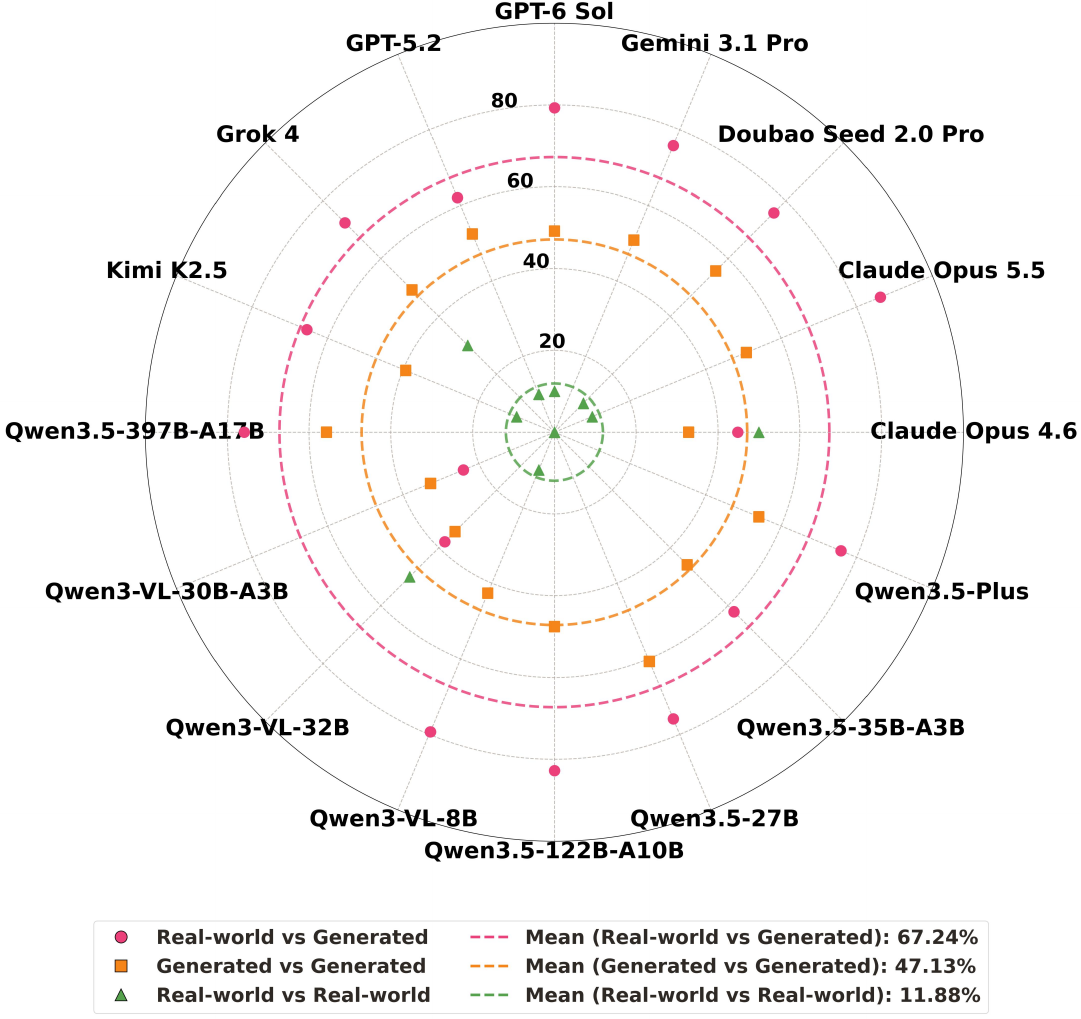} 
    \vfill
    \caption{Accuracy on three types of plausibility comparison.}
    \label{fig:comparison}
  \end{subfigure}
  \hfill
  \begin{subfigure}[c]{0.32\linewidth}
    \centering
    \vfill
    \includegraphics[width=\linewidth]{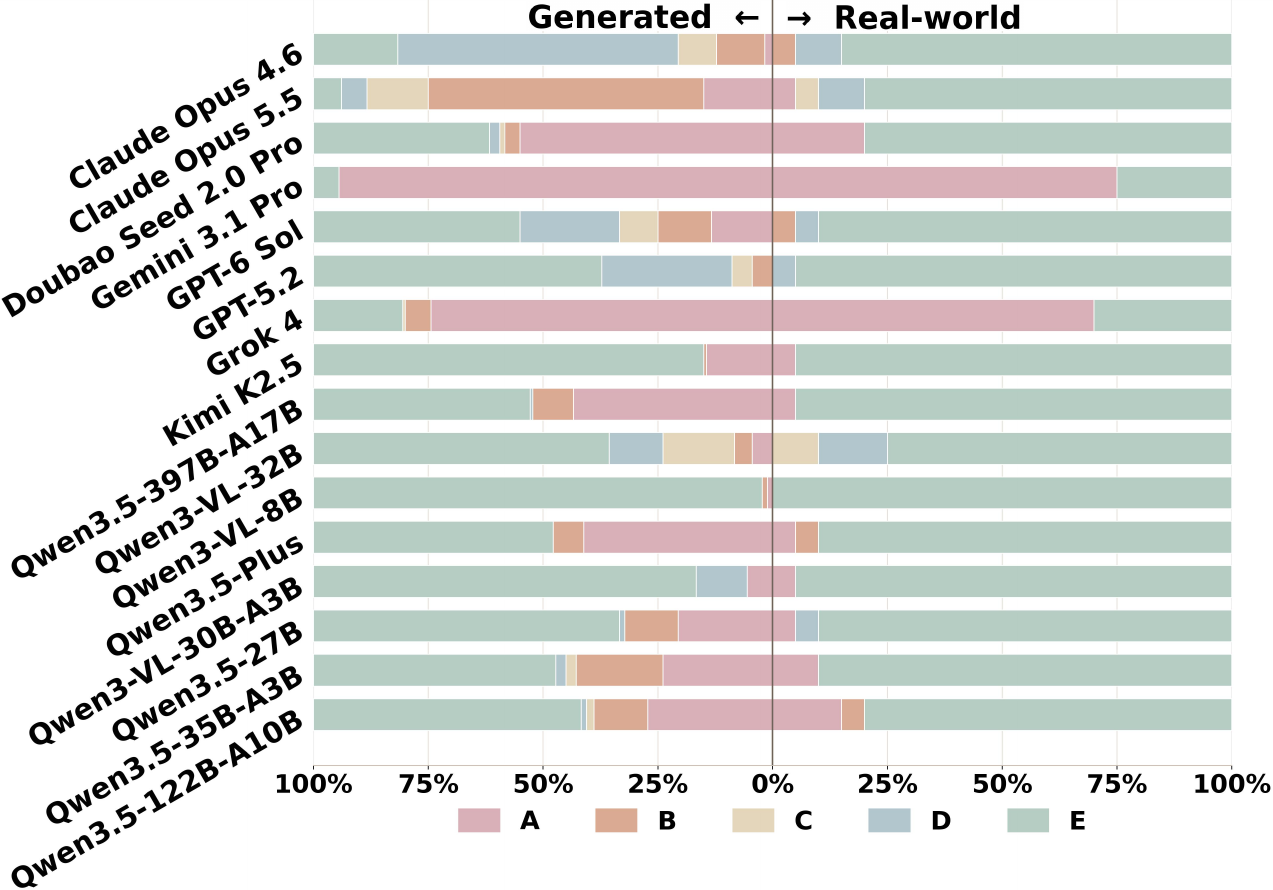}
    \vfill
    \caption{Predicted score distributions for plausibility scoring.}
    \label{fig:score}
  \end{subfigure}

  \caption{Model behavior analyses on PhysVista.}
  \label{fig:total_label}
\end{figure}

\subsection{Overall Benchmark Results}
As shown in Tab.~\ref{tab:except_reasoning}, current VLMs perform relatively better on perception tasks than reasoning and assessment tasks, particularly \emph{Spatial State Perception}, yet remain limited on fine-grained perceptual capabilities such as \emph{Quantitative Scale Estimation} and \emph{Physical Uncertainty Awareness}. 
Tab.~\ref{tab:reasoning} further indicates that \emph{Physical Dynamics Reasoning} is the most challenging stage overall, with especially low accuracy on 
\emph{Physical Dynamics Prediction} 
and \emph{Counterfactual Physical Reasoning}. 
In contrast, results on \emph{Physical Plausibility Assessment} show that models struggle with calibrated, fine-grained plausibility judgment and relative ranking, as evidenced by the low scores on both \emph{Physical Plausibility Scoring} and \emph{Physical Plausibility Comparison} in Tab.~\ref{tab:except_reasoning}.

\vspace{-2mm}
\subsection{Cognitive Gap Analysis} 
 \vspace{-1mm}
Our benchmark is designed to reflect a \emph{seeing--reasoning--assessment} loop. 
The results reveal a clear cognitive gap across the three stages of physical understanding, manifested as substantial \emph{within-model inconsistencies in capability rankings}: models that perform strongly in observable state perception may rank much lower in causal reasoning, while those that excel at physical mechanism reasoning may perform less strongly in physical plausibility assessment.


\textbf{Perception $\rightarrow$ Reasoning gap.}
While GPT-5.2~\cite{gpt} and Kimi K2.5~\cite{team2026kimi} rank third and  forth on \emph{State Perception} (Tab.~\ref{tab:except_reasoning}), respectively, their relative rankings drop substantially on \emph{Dynamics Reasoning} (Tab.~\ref{tab:reasoning}). 
This cross-stage ranking inconsistency suggests that strong performance in observable state perception does not necessarily translate into equally strong causal reasoning. 
Since reasoning about complex dynamic scenarios depends on both perceptual understanding and physical reasoning, these ranking shifts further suggest that the observed limitations cannot be attributed to perceptual capability alone.


\textbf{Reasoning $\rightarrow$ Assessment gap.}
Assessment tasks require global plausibility judgments that are sensitive to severity, rather than mechanism explanations for a certain event. 
While Gemini 3.1 Pro~\cite{Gemini} and Doubao Seed 2.0 Pro~\cite{seed} achieve the third and forth rank on \emph{Physical Dynamics Reasoning} (Tab.~\ref{tab:reasoning}), their performance decreases when evaluated on \emph{Physical Plausibility Assessment} (Tab.~\ref{tab:except_reasoning}), demonstrating a clear capability gap between dynamics reasoning and plausibility assessment. 
This finding highlights a fundamental limitation of current VLMs: although they can sometimes correctly infer the causal mechanisms of violations, they struggle to calibrate physical plausibility along fine-grained levels.

\begin{table*}[t]
\caption{Overall performance on Physical State Perception and Physical Plausibility Assessment tasks. Localization uses mean IoU, scoring uses SRCC/PLCC, and other tasks use accuracy. Avg. Acc. is task-count weighted. Best and second-best results are highlighted in 
{\setlength{\fboxsep}{2pt}\colorbox{brown}{brown}} and 
{\setlength{\fboxsep}{2pt}\colorbox{lightbrown}{light brown}}, 
respectively.}
\label{tab:except_reasoning}
\centering
\resizebox{\textwidth}{!}{%
\begin{tabular}{lcccccccc}
\toprule
Model & Spatial & Camera & Scale 
& Uncertainty & Avg. Acc. & Localization & Score & Comparison  \\
\midrule

GPT-6 Sol~\cite{openai2026gpt6sol}           
& 81.00\% 
& {\setlength{\fboxsep}{2pt}\colorbox{brown}{74.00\%}}
& 35.00\% 
& 60.00\% 
& {\setlength{\fboxsep}{2pt}\colorbox{brown}{68.59\%}}
& {\setlength{\fboxsep}{2pt}\colorbox{lightbrown}{53.03\%}}
& {\setlength{\fboxsep}{2pt}\colorbox{brown}{0.6141}}/{\setlength{\fboxsep}{2pt}\colorbox{lightbrown}{0.5586}}
& 54.00\% 
\\

GPT-5.2~\cite{gpt}             
& {\setlength{\fboxsep}{2pt}\colorbox{brown}{85.00\%}}
& 38.50\% 
& {\setlength{\fboxsep}{2pt}\colorbox{brown}{43.33\%}}
& {\setlength{\fboxsep}{2pt}\colorbox{brown}{76.11\%}}
& 64.06\% 
& 42.09\%
& 0.3328/0.3423
& 51.00\% 
\\

Claude Opus 5.5~\cite{anthropic2026claudeopus55}
& 79.50\% 
& {\setlength{\fboxsep}{2pt}\colorbox{lightbrown}{69.50\%}}
& 35.00\% 
& 56.67\% 
& {\setlength{\fboxsep}{2pt}\colorbox{lightbrown}{65.78\%}}
& {\setlength{\fboxsep}{2pt}\colorbox{brown}{54.87\%}}
& {\setlength{\fboxsep}{2pt}\colorbox{lightbrown}{0.5702}}/{\setlength{\fboxsep}{2pt}\colorbox{brown}{0.5882}}
& {\setlength{\fboxsep}{2pt}\colorbox{lightbrown}{57.00\%}}
\\

Claude Opus 4.6~\cite{claude46}
& 79.00\% 
& 47.50\% 
& 25.00\% 
& 47.22\% 
& 55.15\%
& 46.63\%
& 0.4482/0.4320
& 38.00\%
\\

Gemini 3.1 pro~\cite{Gemini} 
& 74.50\% 
& 54.50\% 
& 31.67\% 
& 53.33\% 
& 58.28\% 
& 45.34\%
& 0.3087/0.3087
& 53.00\% 
\\

Doubao Seed 2.0 Pro~\cite{seed}     
& 77.50\% 
& 59.50\% 
& 30.00\% 
& 50.00\% 
& 59.69\% 
& 47.48\% 
& 0.4018/0.3949
& {\setlength{\fboxsep}{2pt}\colorbox{lightbrown}{57.00\%}}
\\

Grok 4~\cite{Grok4}    
& 72.00\% 
& 29.50\% 
& 25.00\% 
& {\setlength{\fboxsep}{2pt}\colorbox{lightbrown}{71.67\%}}
& 54.22\% 
& 34.61\%
& 0.2853/0.2952
& 54.00\% 
\\

Qwen3.5-Plus~\cite{qwen3.5plus}        
& 82.00\% 
& 48.00\% 
& 26.67\% 
& 38.33\% 
& 53.91\% 
& 45.79\%
& 0.4809/0.4790
& 55.00\% 
\\

Kimi K2.5~\cite{team2026kimi}       
& {\setlength{\fboxsep}{2pt}\colorbox{lightbrown}{84.50\%}}
& 44.00\% 
& {\setlength{\fboxsep}{2pt}\colorbox{lightbrown}{38.33\%}}
& 63.33\% 
& 61.56\% 
& 36.92\%
& 0.2852/0.2849
& 44.00\% 
\\

Qwen3.5-397B-A17B~\cite{qwen3.5-397b-a17b}    
& 83.50\% 
& 51.00\% 
& 28.33\% 
& 41.11\% 
& 56.25\%
& 43.92\%
& 0.4817/0.4788
& 56.00\%
\\

Qwen3.5-122B-A10B~\cite{qwen3.5-122b-a10b}    
& 81.00\% 
& 48.50\% 
& 30.00\% 
& 41.11\% 
& 54.84\%
& 41.05\%
& 0.3593/0.3560
& 53.00\%
\\

Qwen3.5-35B-A3B~\cite{qwen3.5-35b-a3b}    
& 79.50\% 
& 45.00\% 
& 21.67\% 
& 37.22\% 
& 51.41\%
& 42.25\%
& 0.4462/0.4440
& 46.00\%
\\

Qwen3.5-27B~\cite{qwen3.5-27b}    
& 82.50\% 
& 57.00\% 
& 18.33\% 
& 36.67\% 
& 55.63\%
& 42.21\%
& 0.3505/0.3521
& {\setlength{\fboxsep}{2pt}\colorbox{brown}{59.00\%}}
\\

Qwen3-VL-30B-A3B~\cite{Qwen3-VL}    
& 76.00\% 
& 28.50\% 
& 20.00\% 
& 55.56\% 
& 50.16\%
& 39.09\%
& 0.1945/0.2189
& 27.00\%
\\

Qwen3-VL-32B~\cite{Qwen3-VL}    
& 81.00\% 
& 33.00\% 
& 30.00\% 
& 56.67\% 
& 54.38\% 
& 44.61\%
& 0.2105/0.2125
& 37.00\% 
\\

Qwen3-VL-8B~\cite{Qwen3-VL}    
& 75.00\% 
& 22.00\% 
& 23.33\% 
& 34.44\% 
& 42.19\% 
& 48.83\%
& 0.0240/0.0071
& 50.00\% 
\\

\bottomrule
\end{tabular}}
\end{table*}

\begin{table}[t]
\caption{Performance on Physical Dynamics Reasoning. Violation is evaluated by Micro-F1, others
are evaluated by accuracy. Avg. Acc. is task-count weighted. Best and second-best results are highlighted in 
{\setlength{\fboxsep}{2pt}\colorbox{darkblue}{blue}} and 
{\setlength{\fboxsep}{2pt}\colorbox{lightblue}{light blue}}, respectively.}
\label{tab:reasoning}
\centering
\resizebox{\textwidth}{!}{%
\begin{tabular}{lccccccc}
\toprule
Model & Order & Mechanism & Violation & Prediction & Counterfactual & Critique & Avg. Acc.\\
\midrule

GPT-6 Sol~\cite{openai2026gpt6sol}     
& 82.67\% 
& {\setlength{\fboxsep}{2pt}\colorbox{darkblue}{77.60\%}}
& 44.48\% 
& 46.67\% 
& {\setlength{\fboxsep}{2pt}\colorbox{lightblue}{44.54\%}}
& {\setlength{\fboxsep}{2pt}\colorbox{darkblue}{88.00\%}}
& {\setlength{\fboxsep}{2pt}\colorbox{lightblue}{62.68\%}}
\\

GPT-5.2~\cite{gpt}             
& 41.33\% 
& 56.77\% 
& 33.22\% 
& {\setlength{\fboxsep}{2pt}\colorbox{lightblue}{48.15\%}}
& 33.61\% 
& 71.00\% 
& 49.86\% 
\\

Claude Opus 5.5~\cite{anthropic2026claudeopus55}
& {\setlength{\fboxsep}{2pt}\colorbox{lightblue}{90.67\%}}
& {\setlength{\fboxsep}{2pt}\colorbox{lightblue}{77.08\%}}
& {\setlength{\fboxsep}{2pt}\colorbox{darkblue}{51.60\%}}
& {\setlength{\fboxsep}{2pt}\colorbox{darkblue}{58.52\%}}
& {\setlength{\fboxsep}{2pt}\colorbox{darkblue}{49.58\%}}
& {\setlength{\fboxsep}{2pt}\colorbox{lightblue}{87.00\%}}
& {\setlength{\fboxsep}{2pt}\colorbox{darkblue}{67.89\%}}
\\

Claude Opus 4.6~\cite{claude46}
& 52.00\% 
& 68.75\% 
& 41.47\% 
& {\setlength{\fboxsep}{2pt}\colorbox{lightblue}{48.15\%}}
& 36.97\% 
& 82.00\%
& 57.61\%
\\

Gemini 3.1 pro~\cite{Gemini} 
& 66.67\% 
& 74.48\% 
& 40.94\% 
& 40.00\% 
& 38.66\% 
& 76.00\% 
& 60.20\% 
\\

Doubao Seed 2.0 Pro~\cite{seed}     
& {\setlength{\fboxsep}{2pt}\colorbox{darkblue}{91.33\%}}
& 59.90\% 
& 42.09\% 
& 35.56\% 
& 40.34\% 
& 70.00\% 
& 60.06\% 
\\

Grok 4~\cite{Grok4}    
& 46.67\% 
& 47.40\% 
& 45.59\% 
& 22.96\% 
& 27.73\% 
& 56.00\% 
& 40.37\% 
\\

Qwen3.5-Plus~\cite{qwen3.5plus}        
& 62.00\% 
& 68.75\% 
& 45.71\% 
& 39.26\% 
& 41.18\% 
& 78.00\% 
& 58.19\% 
\\

Kimi K2.5~\cite{team2026kimi}       
& 90.00\% 
& 56.25\% 
& 40.17\% 
& 34.81\% 
& 42.02\% 
& 78.00\% 
& 60.06\% 
\\

Qwen3.5-397B-A17B~\cite{qwen3.5-397b-a17b}    
& 68.00\% 
& 69.27\% 
& 43.40\% 
& 36.30\% 
& 37.82\% 
& 73.00\%
& 57.76\%
\\

Qwen3.5-122B-A10B~\cite{qwen3.5-122b-a10b}    
& 62.67\% 
& 62.50\% 
& 48.85\% 
& 31.11\% 
& 32.77\% 
& 73.00\%
& 52.87\%
\\

Qwen3.5-35B-A3B~\cite{qwen3.5-35b-a3b}    
& 55.33\% 
& 59.90\% 
& 48.64\% 
& 40.74\% 
& 28.57\% 
& 70.00\%
& 51.29\%
\\

Qwen3.5-27B~\cite{qwen3.5-27b}    
& 62.67\% 
& 64.06\% 
& {\setlength{\fboxsep}{2pt}\colorbox{lightblue}{49.82\%}}
& 30.37\% 
& 36.13\% 
& 72.00\%
& 53.59\%
\\

Qwen3-VL-30B-A3B~\cite{Qwen3-VL}    
& 60.67\% 
& 52.08\% 
& 37.98\% 
& 40.00\% 
& 31.93\% 
& 78.00\%
& 51.87\%
\\

Qwen3-VL-32B~\cite{Qwen3-VL}    
& 70.00\% 
& 61.46\% 
& 40.20\% 
& 45.93\% 
& 26.05\% 
& 71.00\% 
& 55.60\%
\\

Qwen3-VL-8B~\cite{Qwen3-VL}    
& 57.33\% 
& 54.69\% 
& 41.31\% 
& 28.89\% 
& 26.89\% 
& 63.00\% 
& 46.70\% 
\\

\bottomrule
\end{tabular}
}
\end{table}

\subsection{Subtask Difficulty and Failure Modes in Reasoning}
\vspace{-1mm}
Tab.~\ref{tab:reasoning} highlights strong heterogeneity in reasoning difficulty.

\textbf{Temporal reasoning is easier than causal reasoning.}
As for \emph{Temporal Order Reconstruction}, models generally perform better on this task than on causal reasoning tasks, suggesting that temporal ordering can often be solved using basic physical commonsense with temporally coherent, whereas causal tasks require more complex reasoning with explicit physical modeling. 


\textbf{Prediction and Counterfactual reasoning remain difficult.}
\emph{Physical Dynamics Prediction} and \emph{Counterfactual Physical Reasoning} rely more heavily on temporal information analysis than other causal reasoning tasks, as models must infer how the observed dynamics evolve over time rather than only identifying causal mechanisms. 
This makes it difficult for current models to simultaneously maintain event-level correctness and fine-grained scale-level precision. 
Although both tasks involve temporal reasoning and causal understanding, they differ in their reasoning requirements: \emph{Physical Dynamics Prediction} can often be solved through temporal extrapolation combined with causal reasoning about the observed motion, whereas \emph{Counterfactual Physical Reasoning} additionally requires reasoning under hypothetical interventions and implicitly simulating the resulting dynamics. 
This difference in reasoning demands likely accounts for the performance discrepancy between the two tasks within the same model.

\vspace{-1.8mm}
\subsection{Real-world vs. Generated Content Analysis}
\vspace{-1.5mm}
To analyze model behavior across data sources, we compare performance on real-world and generated videos across multiple tasks (Fig.~\ref{fig:gap}). Overall, models perform better on real-world videos, revealing a systematic generalization gap.

\textbf{Generated content amplifies subtle failure modes.}
Across most tasks, accuracy on real-world videos is higher than on generated videos. 
While basic spatial perception and temporal reasoning remain relatively stable, showing only small decreases on generated content, the performance gap becomes much larger for \emph{Physical Uncertainty Awareness} and \emph{Physical Dynamics Prediction}. 
These tasks require abstract perception or fine-grained physical judgment, making them particularly sensitive to subtle physical inconsistencies that often appear in generated videos. 
This suggests that current models rely heavily on visually and physically consistent patterns commonly observed in real-world data. 
When subtle violations appear, models often fail to internally reconcile these inconsistencies during the reasoning process, revealing a lack of deeper physical reasoning capability. 

\textbf{Not all tasks become harder on generated videos.}
An interesting exception appears in \emph{Quantitative Scale Estimation}, where performance on generated videos slightly exceeds that on real-world videos. 
One possible explanation is that generated videos may present cleaner geometric layouts or more controlled object configurations, making relative scale relationships easier to infer compared to real-world scenes with cluttered backgrounds and measurement ambiguity. 

\vspace{-1mm}
\subsection{Assessment Task Analysis}
\label{subsec:assessment_analysis}
\vspace{-0.8mm}
Physical Plausibility Assessment consists of two tasks: 
\emph{Physical Plausibility Scoring} (single-video, 1--5 rating) and 
\emph{Physical Plausibility Comparison} (pairwise ranking). 
We analyze models' scoring calibration and distribution patterns in the scoring task, as well as their decision behavior across different pair types in the comparison task. 

\textbf{Physical Plausibility Scoring: Calibration vs. Score Collapse}
Fig.~\ref{fig:score} visualizes the distribution of predicted scores for generated and real-world videos. 
We observe the critical failure mode:
\textbf{1) Score collapse in plausibility scoring.}
We observe a consistent \emph{score collapse} phenomenon in several models (\textit{e.g.}, Qwen3-VL-8B and Kimi K2.5), where a disproportionately large fraction of predictions concentrate on the highest score level for \emph{both} generated and real-world videos. 
This phenomenon suggests that many models are unable to effectively differentiate between varying levels of physical plausibility, but instead default to a dominant score when clear physical violations are not confidently identified. 
Consequently, correct predictions on real-world videos may arise merely because these samples are labeled with the highest score—the default prediction for many models—rather than reflecting accurate plausibility estimation. 
Notably, the collapse is model-dependent: some model shift the default score to the lowest level (\textit{e.g.}, Gemini 3.1 Pro and Grok 4), performance on real-world videos deteriorates dramatically. 
This difference likely stems from variations in their inherent priors toward certain score tokens. 
\textbf{2) Competent scoring requires distributional separation.}
Models that exhibit genuine scoring ability show distinct distributions between generated and real-world videos: real-world predictions concentrate on high plausibility consistently, while generated predictions spread across lower levels. 
This separation indicates sensitivity to graded realism and better calibration to violation severity. 

\textbf{Physical Plausibility Comparison: Varying Difficulty Across Pair Types}
The comparison task includes three pair types: \emph{Real-world vs.\ Generated}, \emph{Generated vs.\ Generated}, and \emph{Real-world vs.\ Real-world}, with results shown in Fig.~\ref{fig:comparison}. 
\textbf{1) Real-world vs.\ Generated is easiest.} This setting exhibits clear physical contrast, leading to the highest accuracy and mainly testing the ability to reject obvious implausibility. 
\textbf{2) Generated vs.\ Generated is substantially harder.} Accuracy drops when comparing two generated videos with similar semantics, requiring fine-grained evaluation of physical coherence and ranking of plausibility among imperfect samples. 
\textbf{3) Real-world vs.\ Real-world exposes over-confident preference.} When both videos are physically valid, models still tend to force a choice, leading to low accuracy and indicating limited ability to recognize equivalence. This highlights the need for models to explicitly handle the ``equal'' case.

\vspace{-1mm}
\section{Conclusion}
\vspace{-2.6mm}
In this work, we introduce PhysVista, a benchmark for evaluating physical intelligence in Vision-Language Models through a closed cognitive loop of Physical State Perception, Physical Dynamics Reasoning, and Physical Plausibility Assessment. Unlike prior benchmarks that focus on isolated abilities, PhysVista provides a unified evaluation across real-world and AI-generated videos. Extensive experiments show that although modern VLMs exhibit strong semantic perception, their physical intelligence remains incomplete across the full cognitive loop. Nevertheless, PhysVista still has limitations in evaluation granularity and the depth and breadth of its question coverage, leaving room for more fine-grained and comprehensive evaluations of physical intelligence. We hope this diagnostic framework will facilitate future research on physically grounded vision-language systems.

\vspace{-1mm}
\begin{ack}
\vspace{-1.8mm}
This work was supported in part by the National Natural Science Foundation of China (NSFC) under Grants 62371434 and U25B2010, the Fundamental Research Funds for the Central Universities (No. WK2100250064), and ZGCA Project C20250302.
\end{ack}

\clearpage

\bibliographystyle{abbrv}
\bibliography{ref}

\clearpage
\appendix
\section{Technical appendices and supplementary material}

\subsection{Definition of Camera Motions}
\label{sec:intro}
We incorporate 10 camera motion types in Camera Motion Recognition task, the definition of each type is illustrated in Tab.~\ref{tab:camera_motions}. 

\vspace{-3mm}
\begin{table}[H]
  \centering
  \caption{Definition of Camera Motions.}
  \label{tab:camera_motions}
  \renewcommand{\arraystretch}{1.4} 
  \begin{tabularx}{\textwidth}{@{}l X@{}}
    \toprule
    \textbf{Camera Motion} & \textbf{Definition} \\
    \midrule
    Arc Left/Right & Horizontal circular movement around a subject, keeping it centered. \\
    Pan Left/Right & Horizontal rotation of the camera on a fixed axis. \\
    Tilt Up/Down & Vertical rotation of the camera on a fixed axis. \\
    Translate Up/Down & Vertical displacement while adjusting pitch to maintain focus. \\
    Static & The camera remains stationary with no movement or rotation. \\
    Zoom Out & Decreasing the lens focal length to widen the field of view. \\
    \bottomrule
  \end{tabularx}
\end{table}

\subsection{Benchmark Statistics}
\label{statistics}
\begin{figure}[H]
  \centering
  \begin{subfigure}{0.3\linewidth} 
    \includegraphics[width=\linewidth]{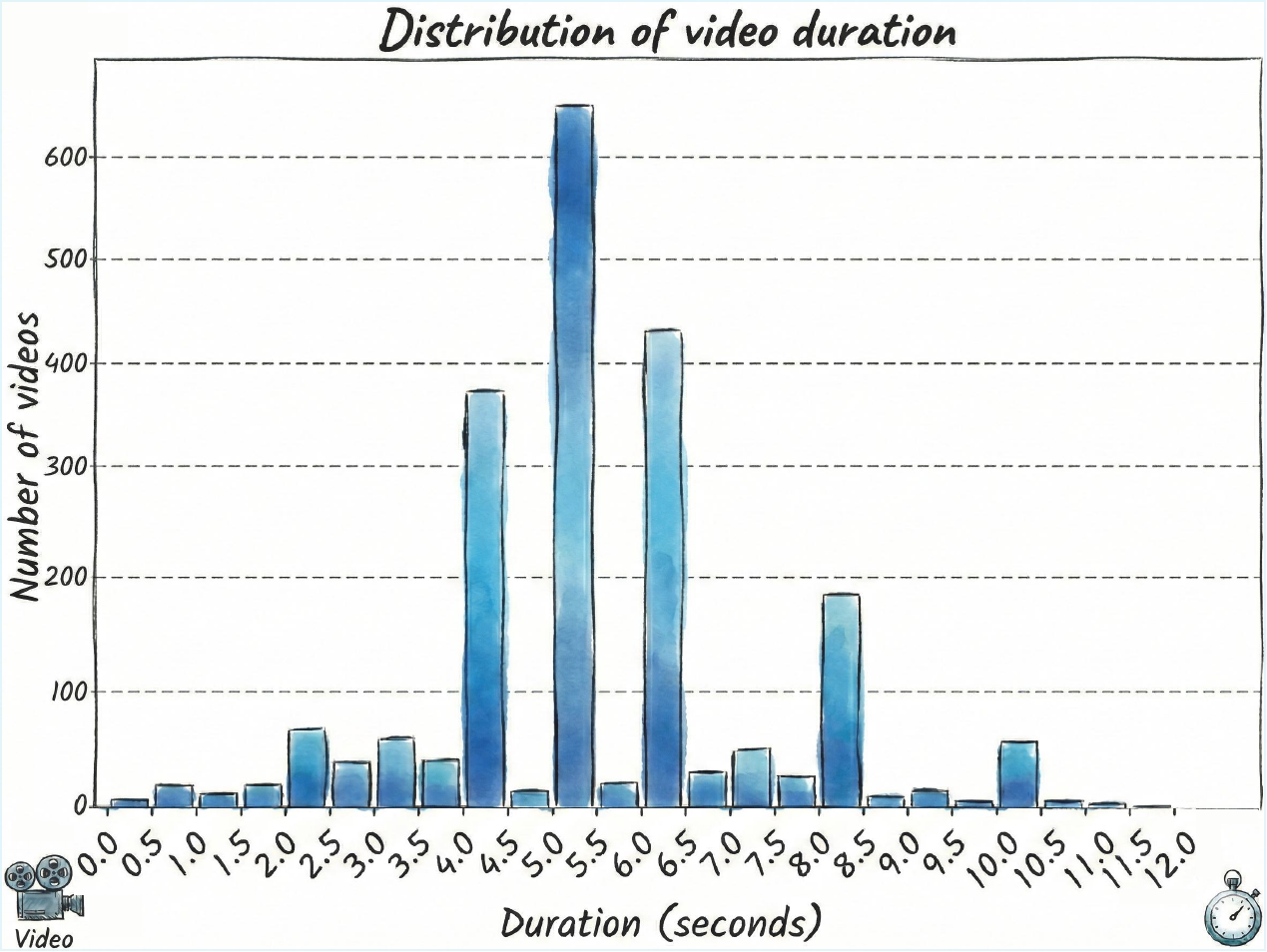}
    \caption{Distribution of video durations.}
    \label{fig:duration}
  \end{subfigure}
  \hfill 
  \begin{subfigure}{0.3\linewidth}
    \includegraphics[width=\linewidth]{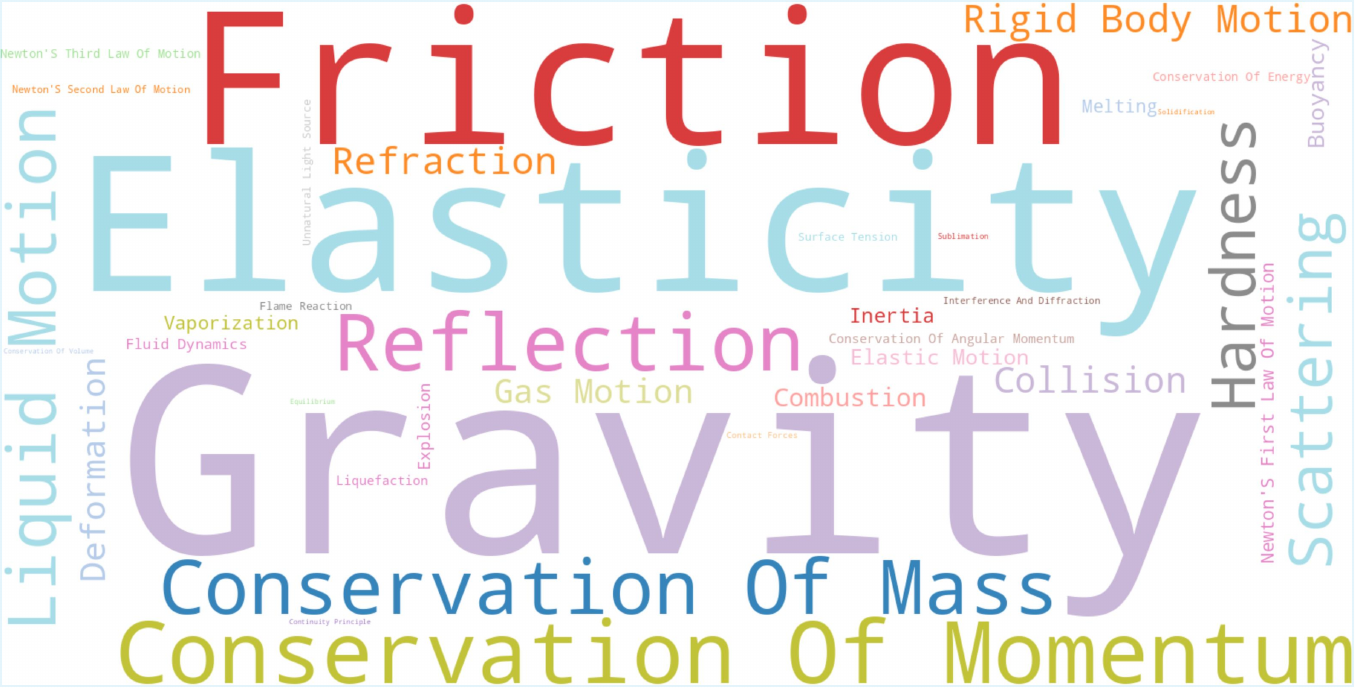}
    \caption{Word cloud of physical principles evaluated in the benchmark.}
    \label{fig:physics_wordcloud}
  \end{subfigure}
  \hfill
  \begin{subfigure}{0.3\linewidth}
    \includegraphics[width=\linewidth]{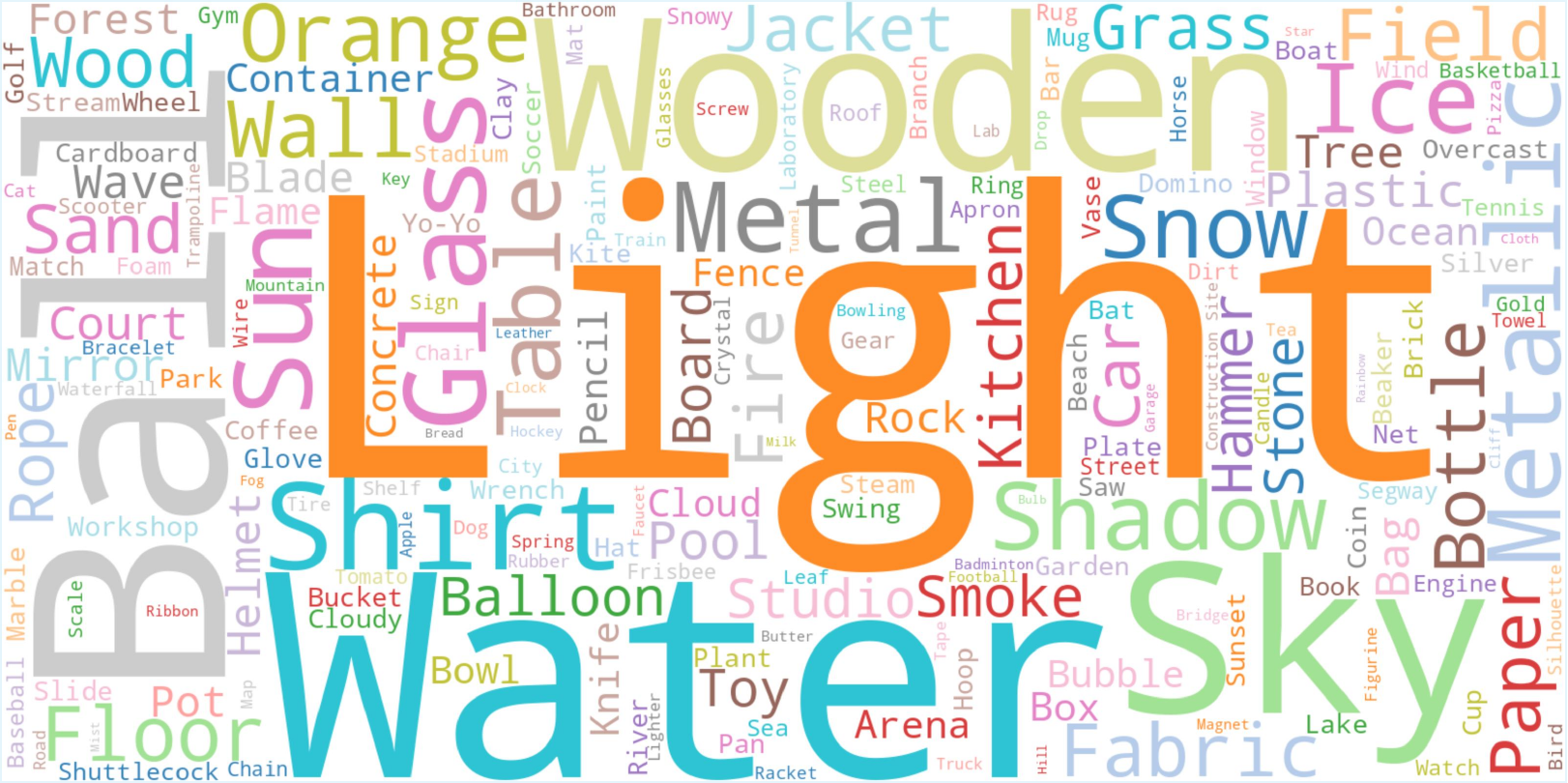}
    \caption{Word cloud of video captions summarizing the covered scenarios.}
    \label{fig:scene_wrodcloud}
  \end{subfigure}
  \caption{Data statistics of PhysVista.}
  \vspace{-6.5mm}
  \label{fig:total_label}
\end{figure}

The statistics of video duration is illustrated in Fig.~\ref{fig:duration}. 
The word cloud of physical principles from the video is demonstrated in Fig.~\ref{fig:physics_wordcloud}. 
Fig.~\ref{fig:scene_wrodcloud} shows a word cloud of the video captions, reflecting the content present in the scenes. 
We also shown the number of each tasks in Tab.~\ref{tab:number1} and Tab.~\ref{tab:number2}. 

\begin{table}[H]
\caption{Number of samples for each task in Physical State Perception.}
\label{tab:number1}
\centering
\resizebox{0.6\textwidth}{!}{%
\begin{tabular}{ccccc}
\toprule
Spatial & Localization & Camera & Scale 
& Uncertainty \\
\midrule
 200
& 200
& 200
& 60
& 180

\\

\bottomrule
\end{tabular}
}
\end{table}

\begin{table}[H]
\caption{Number of samples for each task in Physical Dynamics Reasoning and Physical Plausibility Assessment.}
\label{tab:number2}
\centering
\resizebox{\textwidth}{!}{%
\begin{tabular}{cccccccc}
\toprule
 Order & Mechanism & Violation & Prediction & Counterfactual & Critique & Score & Comparison\\
\midrule
 150
& 192
& 224
& 135
& 119
& 100
& 200
& 100

\\
\bottomrule
\end{tabular}
}
\end{table}

\subsection{Human Annotation, Verification, and Evaluation}
In data generation stage, we incorporate six human annotators for data annotation and verification. 

\begin{figure}[H]
  \centering
\includegraphics[width=0.9\linewidth]{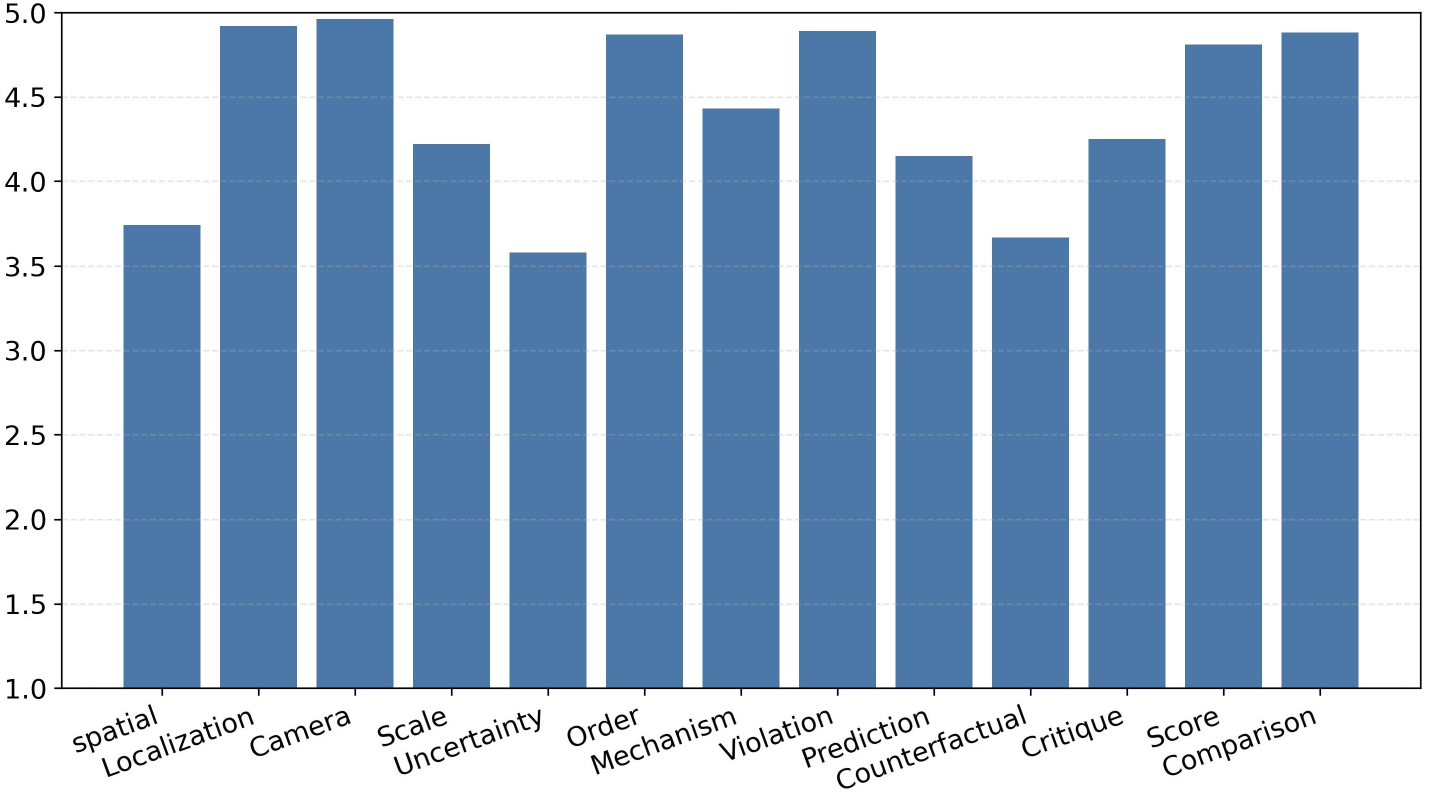}
  \caption{The results of linguistic neutrality evaluation.
  }
  \label{fig:quality_linguistic}
\end{figure}

\begin{figure}[H]
  \centering
\includegraphics[width=0.9\linewidth]{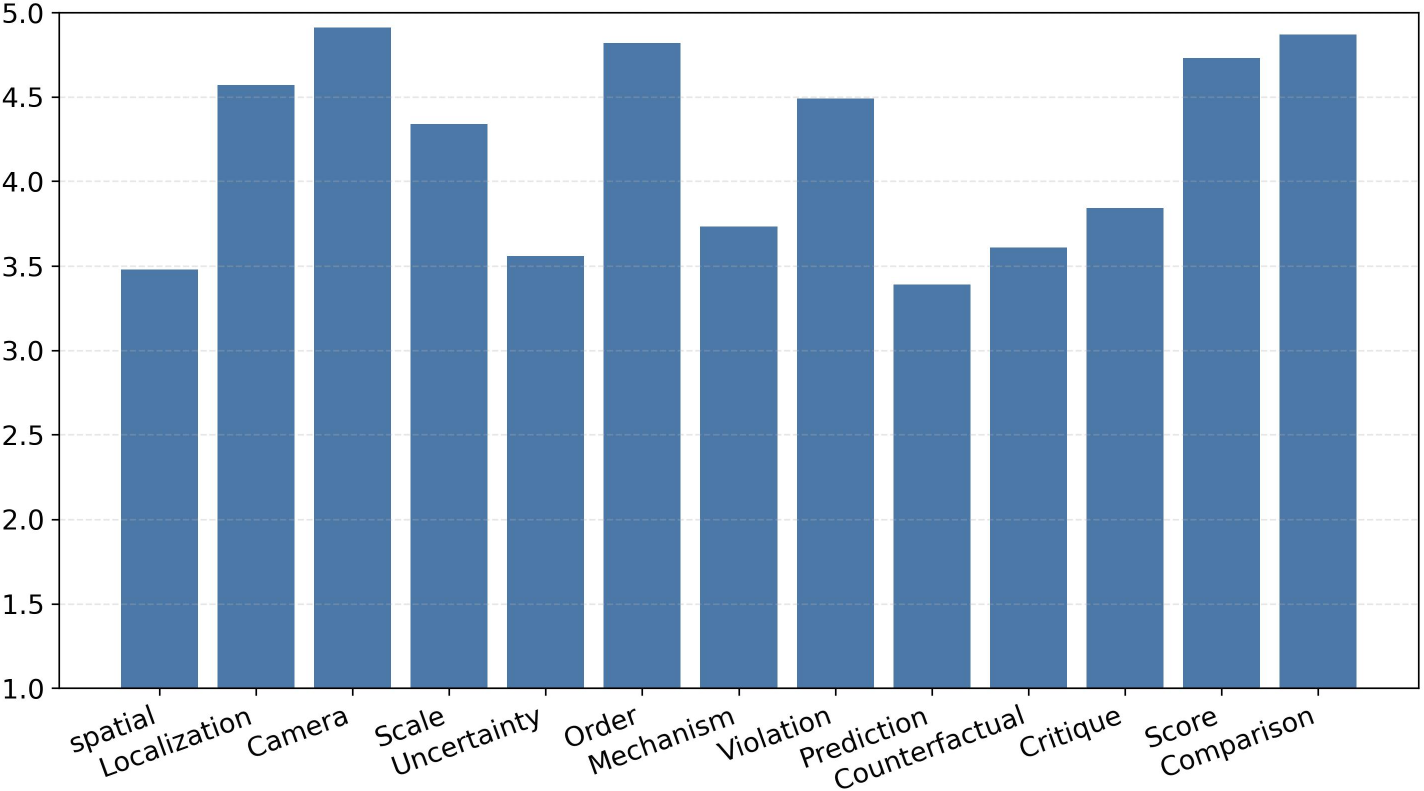}
  \caption{The results of question-answer precision evaluation.
  }
  \label{fig:quality_question}
\end{figure}

\begin{figure}[H]
  \centering
\includegraphics[width=0.9\linewidth]{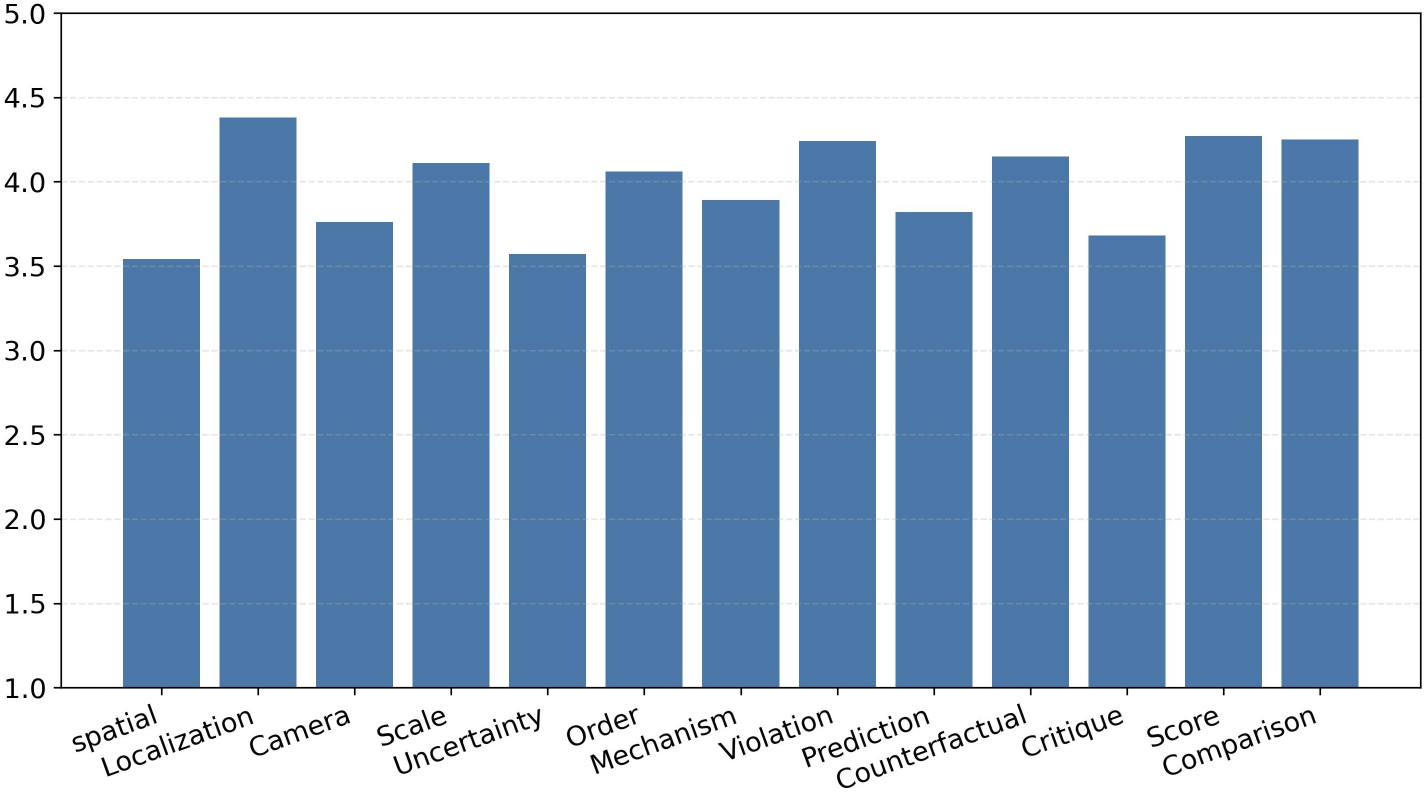}
  \caption{The results of physical reasoning validity evaluation.
  }
  \label{fig:quality_physical}
\end{figure}

In data evaluation stage, we recruit five independent human evaluators to rate the 40\% of the benchmark dataset on a scale from 1 to 5 in terms of linguistic neutrality, question-answer precision, and physical reasoning validity. The results are presented in Fig.~\ref{fig:quality_linguistic}, Fig.~\ref{fig:quality_question} and Fig.~\ref{fig:quality_physical}.

\subsection{More Results}
For the Physical Principle Violation Reasoning task, we additionally report exact-match accuracy, option-level Micro-F1, and principle-type Macro-F1 in Tab.~\ref{tab:vio}. Option-level Micro-F1 aggregates true positives, false positives, and false negatives over all A/B/C/D option predictions, measuring the overall multi-label prediction performance. Principle-type Macro-F1 first maps each selected option to its corresponding physical-principle category and then averages F1 scores across principle types, reflecting whether models perform consistently across different physical laws. 
This task requires mapping observed dynamics to principle-level constraints and often involves multiple simultaneous violations. 
Most models achieve relatively low performance on this task, suggesting that many VLMs lack reliable principle-grounded abstractions even when violations are visually salient. 

\begin{table}[H]
\caption{Performance on Physical Principle Violation Reasoning task. Best and second-best results are highlighted in 
{\setlength{\fboxsep}{2pt}\colorbox{pink!70}{red}} and 
{\setlength{\fboxsep}{2pt}\colorbox{pink!30}{light red}}, respectively.}
\label{tab:vio}
\centering
\resizebox{0.9\textwidth}{!}{%
\begin{tabular}{lccc}
\toprule
Model & Exact-match Accuracy & Macro-F1 & Micro-F1\\
\midrule

GPT-6 Sol~\cite{openai2026gpt6sol}        
& {\setlength{\fboxsep}{2pt}\colorbox{pink!30}{20.54\%}}
& 46.60\% 
& 44.48\%
\\

GPT-5.2~\cite{gpt}             
& 14.73\% 
& 35.93\% 
& 33.22\%
\\

Claude Opus 5.5~\cite{anthropic2026claudeopus55}
& 15.63\% 
& {\setlength{\fboxsep}{2pt}\colorbox{pink!70}{60.41\%}}
& {\setlength{\fboxsep}{2pt}\colorbox{pink!70}{51.60\%}}
\\

Claude Opus 4.6~\cite{claude46}
& 12.05\% 
& 48.41\% 
& 41.47\%
\\

Gemini 3.1 pro~\cite{Gemini} 
& 15.62\% 
& 42.77\% 
& 40.94\%
\\

Doubao Seed 2.0 Pro~\cite{seed}     
& 14.29\% 
& 49.12\% 
& 42.09\%
\\

Grok 4~\cite{Grok4}    
& 11.61\% 
& 50.03\%
& 45.59\%
\\

Qwen3.5-Plus~\cite{qwen3.5plus}        
& 12.05\% 
& 47.21\%
& 45.71\%
\\

Kimi K2.5~\cite{team2026kimi}       
& 10.27\% 
& 40.98\%
& 40.17\%
\\

Qwen3.5-397B-A17B~\cite{qwen3.5-397b-a17b}    
& 7.14\% 
& 45.56\%
& 43.40\%
\\

Qwen3.5-122B-A10B~\cite{qwen3.5-122b-a10b}    
& 10.71\% 
& 53.79\%
& 48.85\%
\\

Qwen3.5-35B-A3B~\cite{qwen3.5-35b-a3b}    
& 10.71\% 
& 54.26\%
& 48.64\%
\\

Qwen3.5-27B~\cite{qwen3.5-27b}    
& 8.48\% 
& {\setlength{\fboxsep}{2pt}\colorbox{pink!30}{58.59\%}}
& {\setlength{\fboxsep}{2pt}\colorbox{pink!30}{49.82\%}}
\\

Qwen3-VL-30B-A3B~\cite{Qwen3-VL}    
& 5.36\% 
& 34.90\%
& 37.98\%
\\

Qwen3-VL-32B~\cite{Qwen3-VL}    
& 16.07\% 
& 41.16\%
& 40.20\%
\\

Qwen3-VL-8B~\cite{Qwen3-VL}    
& {\setlength{\fboxsep}{2pt}\colorbox{pink!70}{20.98\%}}
& 31.66\%
& 41.31\%
\\

\bottomrule
\end{tabular}
}
\end{table}

\subsection{Task Illustrations}
\begin{figure}[tb]
  \centering
\includegraphics[width=1\linewidth]{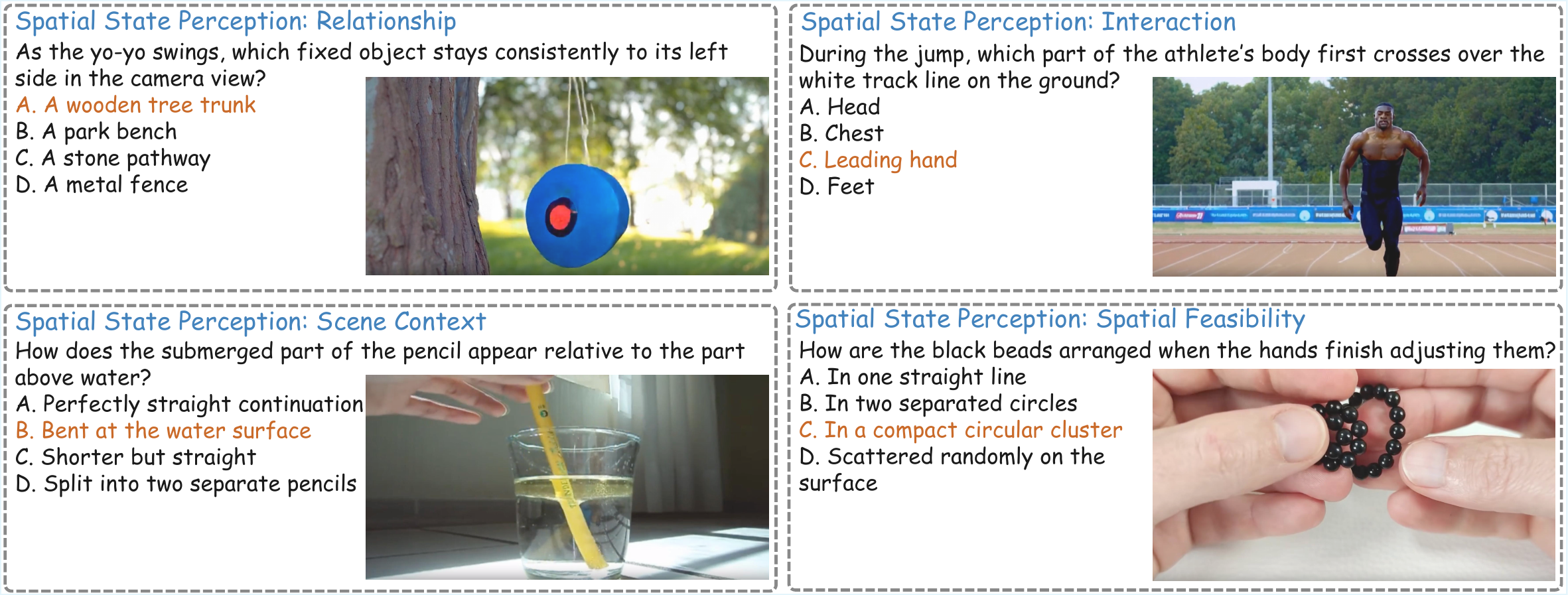}
  \caption{Spatial State Perception Tasks.
  }
  \label{fig:sp_spatial}
\end{figure}

\begin{figure}[tb]
  \centering
\includegraphics[width=1\linewidth]{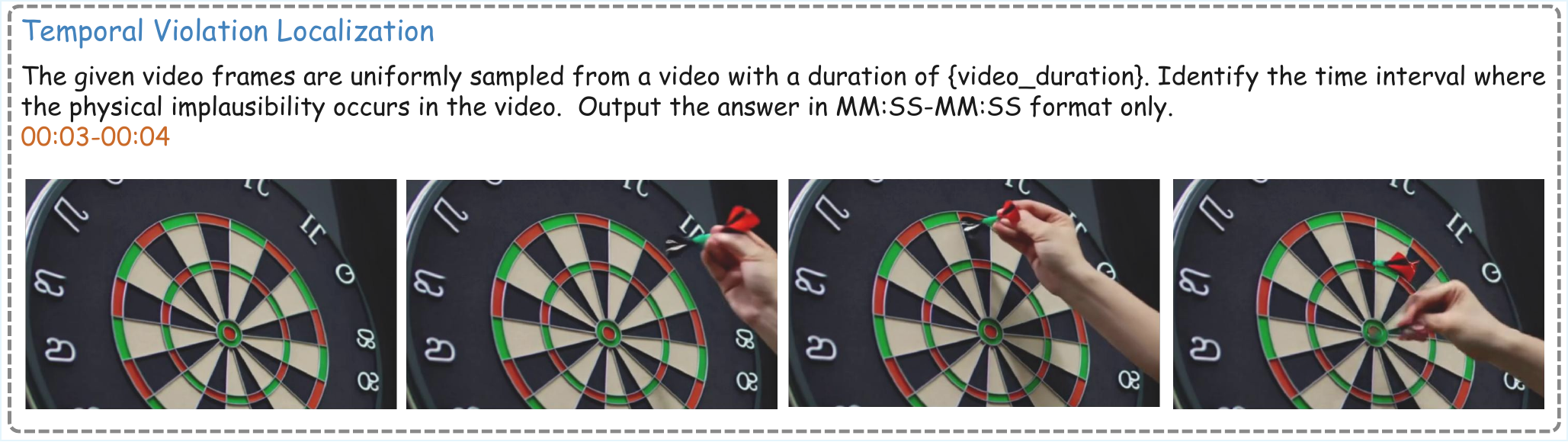}
  \caption{Temporal Violation Localization Task.
  }
  \label{fig:sp_localization}
\end{figure}

\begin{figure}[tb]
  \centering
\includegraphics[width=1\linewidth]{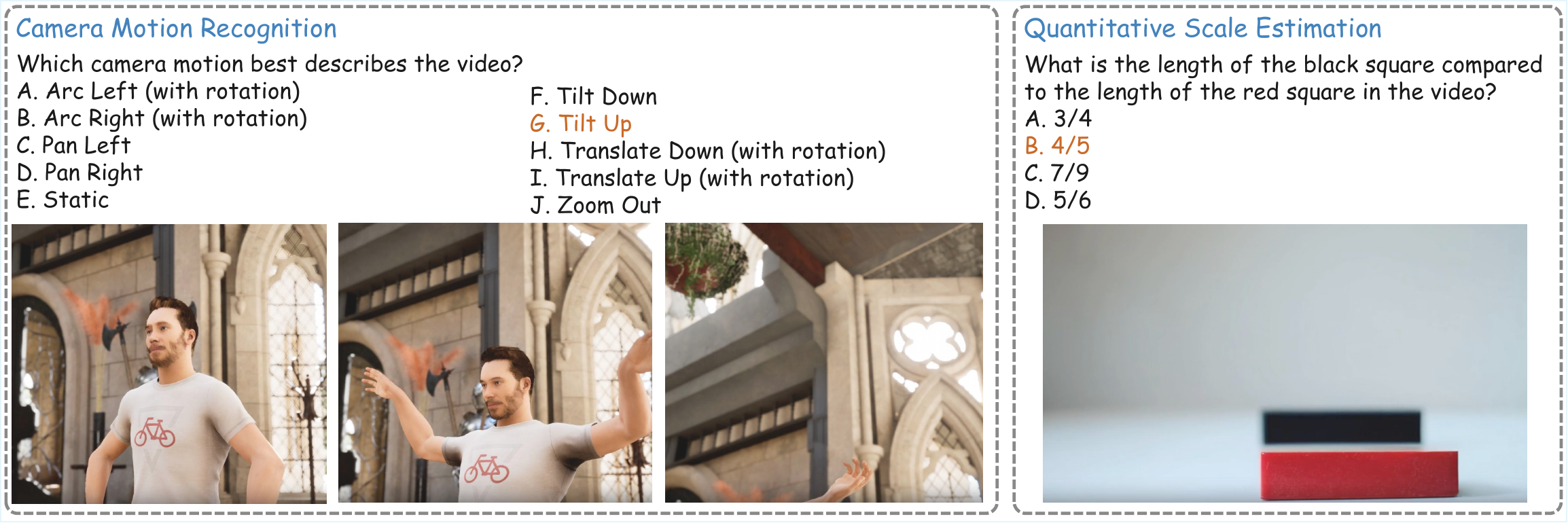}
  \caption{Camera Motion Recognition and Quantitative Scale Estimation Task.
  }
  \label{fig:sp_camera}
\end{figure}

\begin{figure}[tb]
  \centering
\includegraphics[width=1\linewidth]{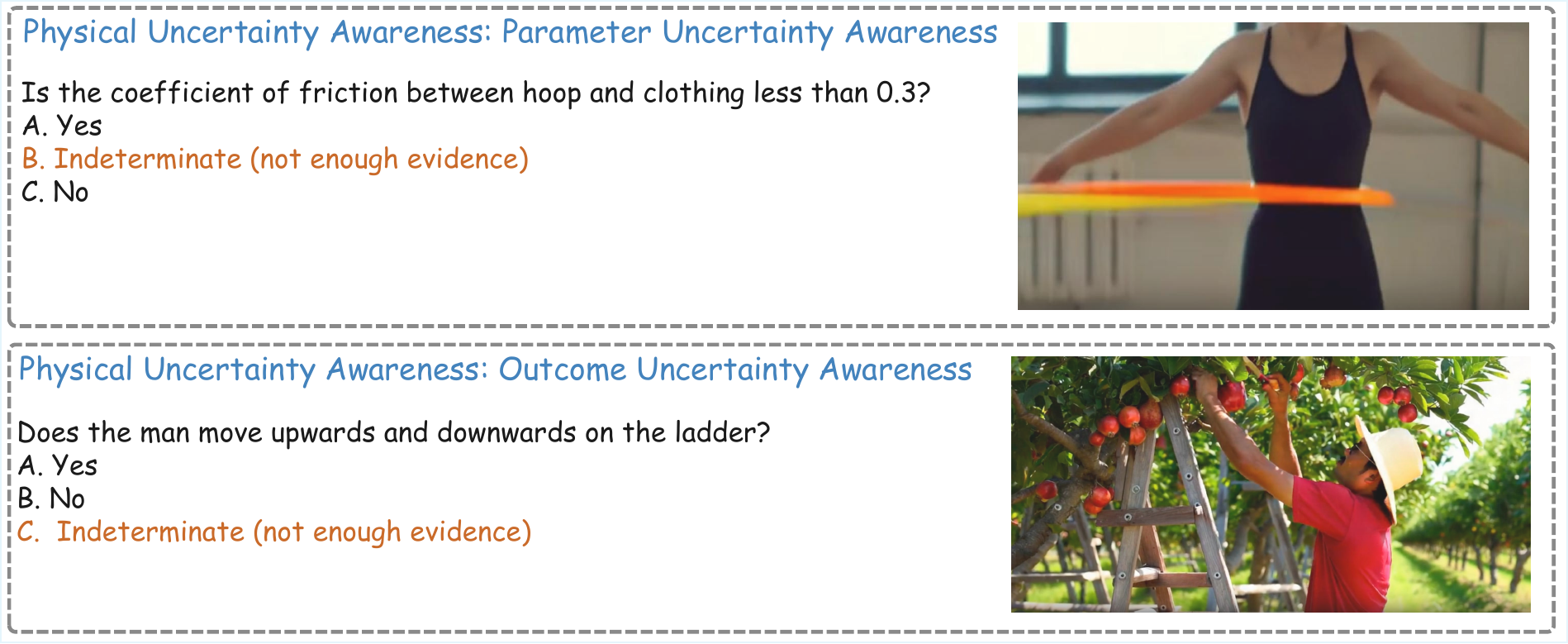}
  \caption{Physical Uncertainty Awareness Task.
  }
  \label{fig:sp_uncertainty}
\end{figure}

\begin{figure}[tb]
  \centering
\includegraphics[width=1\linewidth]{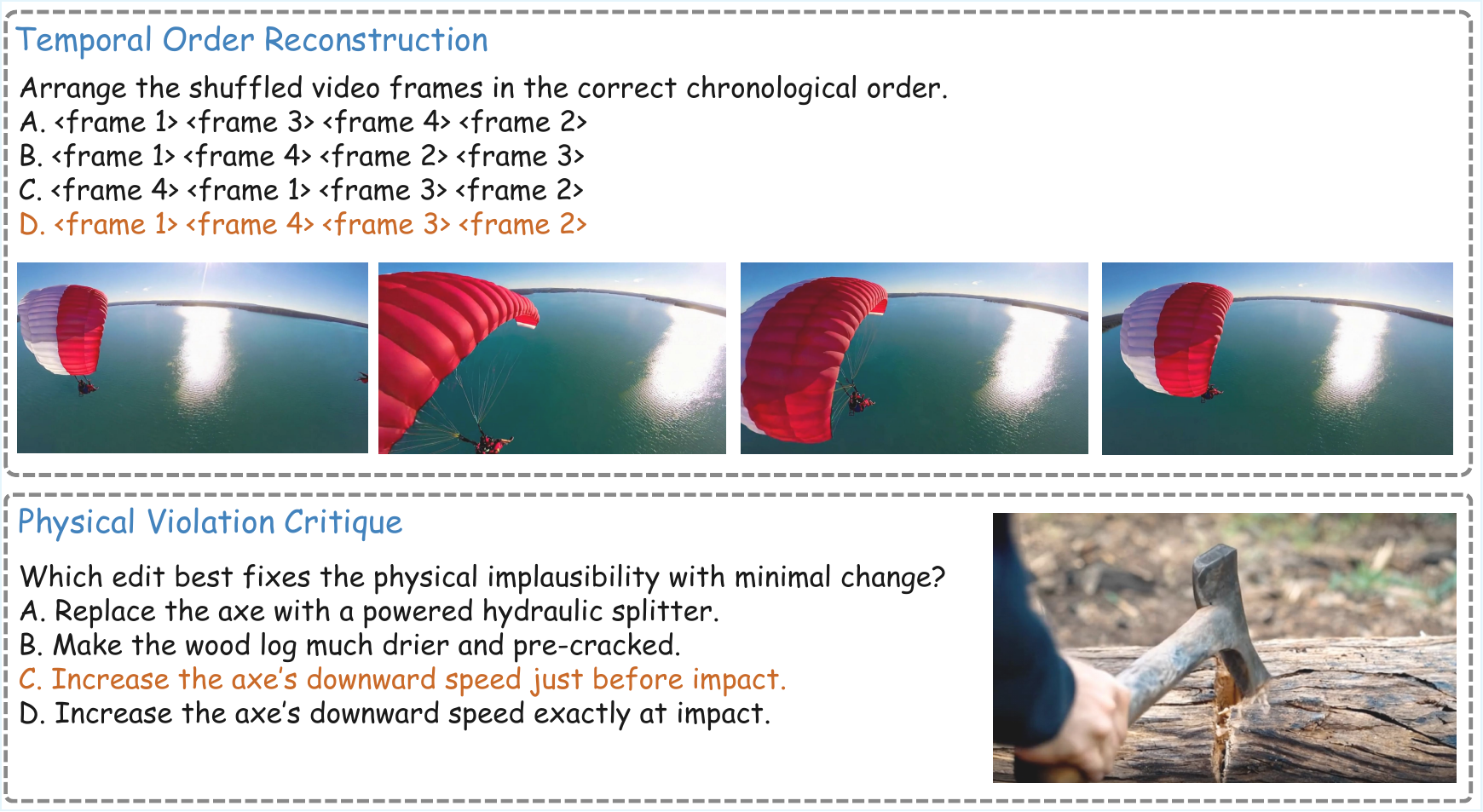}
  \caption{Temporal Order Reconstruction and Physical Violation Critique Task.
  }
  \label{fig:sp_order}
\end{figure}

\begin{figure}[tb]
  \centering
\includegraphics[width=1\linewidth]{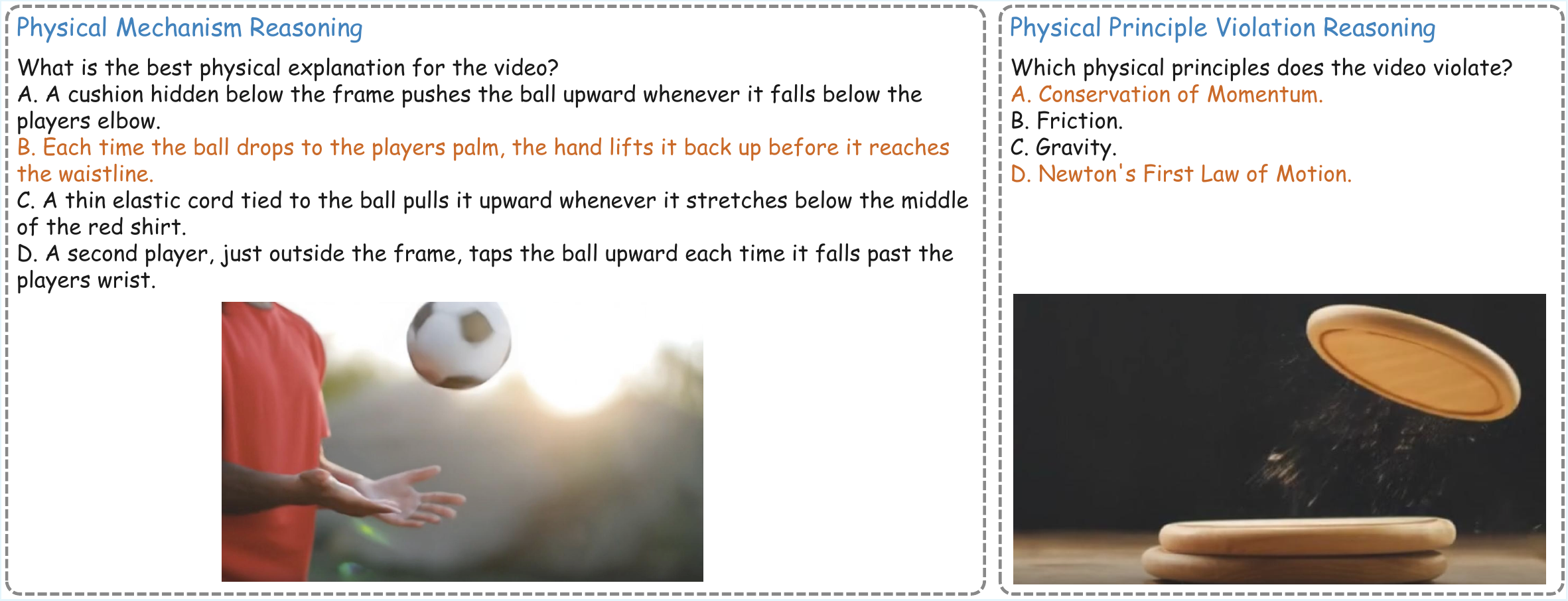}
  \caption{Physical Mechanism Reasoning and Physical Principle Violation Reasoning Task.
  }
  \label{fig:sp_mechanism}
\end{figure}

\begin{figure}[tb]
  \centering
\includegraphics[width=1\linewidth]{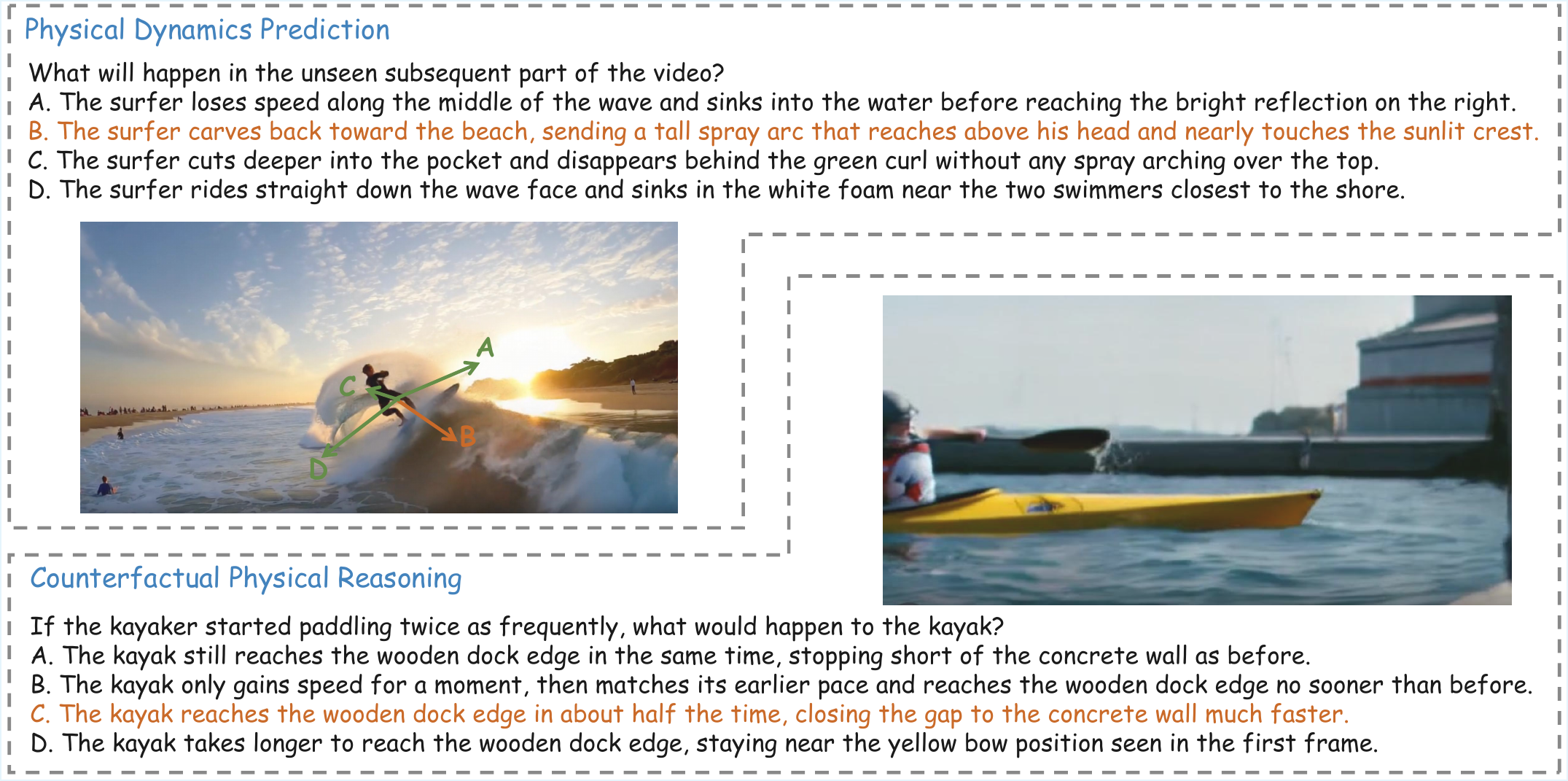}
  \caption{Physical Dynamics Prediction and Counterfactual Physical Reasoning Task.
  }
  \label{fig:sp_prediction}
\end{figure}

\begin{figure}[tb]
  \centering
\includegraphics[width=1\linewidth]{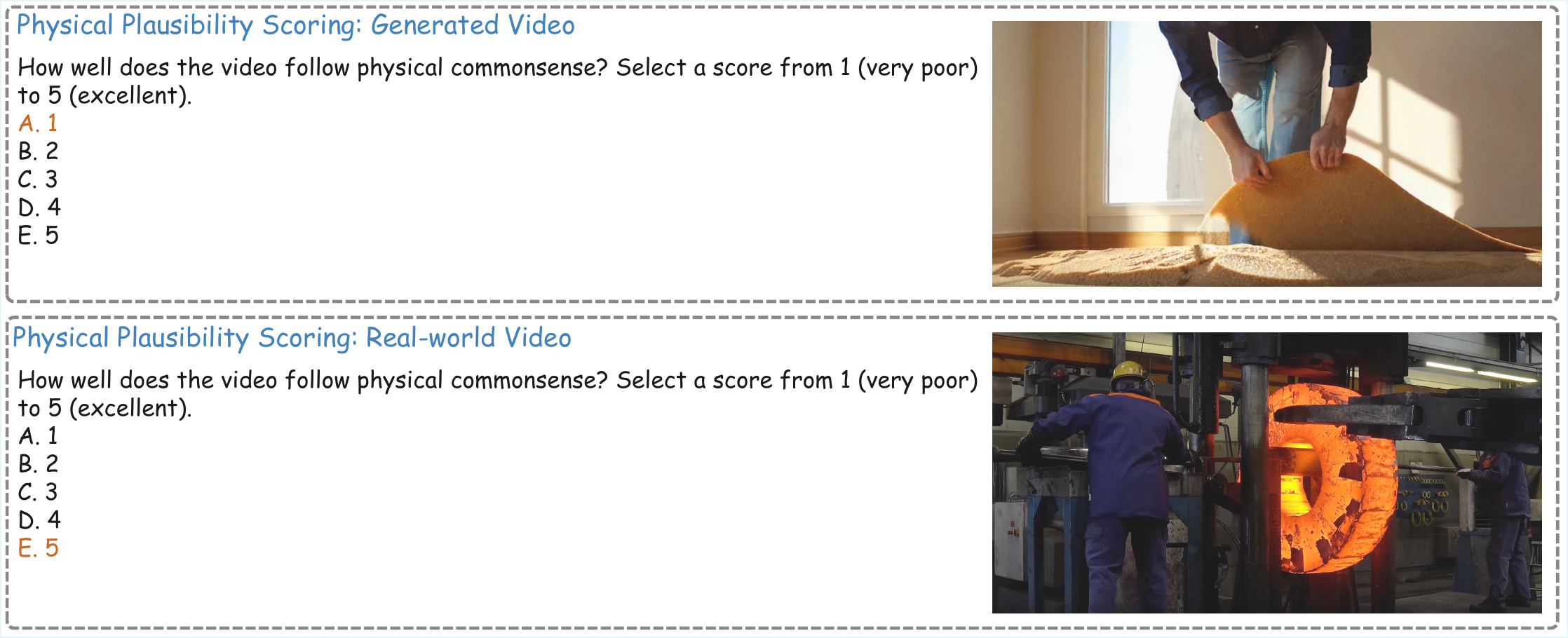}
  \caption{Physical Plausibility Scoring Task.
  }
  \label{fig:sp_score}
\end{figure}

\begin{figure}[tb]
  \centering
\includegraphics[width=1\linewidth]{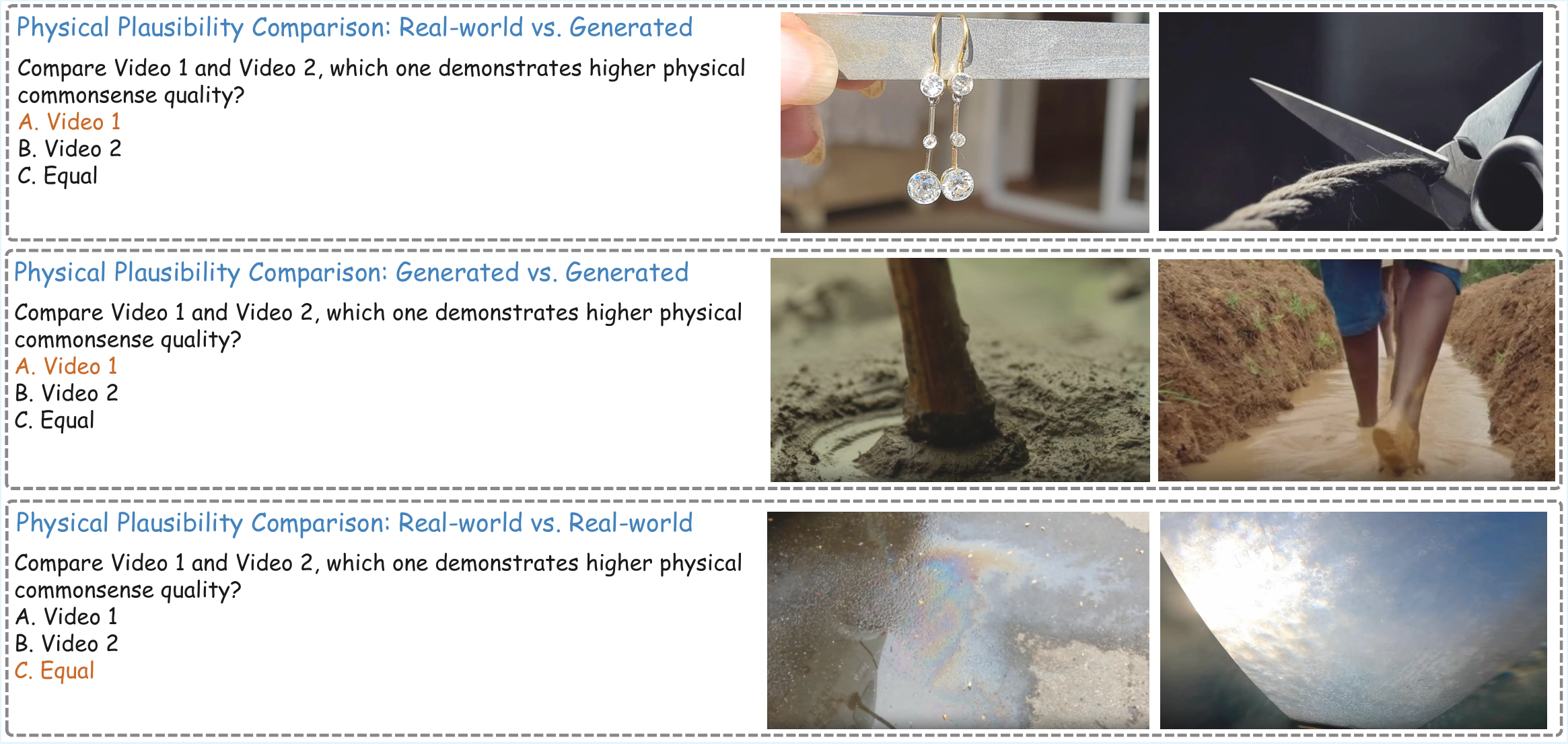}
  \caption{Physical Plausibility Comparison Task.
  }
  \label{fig:sp_compare}
\end{figure}

Our benchmark includes tasks forming a closed loop: Physical State Perception, Physical Dynamics Reasoning, and Physical Plausibility Assessment. 
Examples for these tasks are illustrated in Fig.~\ref{fig:sp_spatial}, Fig.~\ref{fig:sp_localization}, 
Fig.~\ref{fig:sp_camera},
Fig.~\ref{fig:sp_uncertainty},
Fig.~\ref{fig:sp_order},
Fig.~\ref{fig:sp_mechanism},
Fig.~\ref{fig:sp_prediction},
Fig.~\ref{fig:sp_score},
and Fig.~\ref{fig:sp_compare}.

\clearpage

\end{document}